\pdfoutput=1
\documentclass[10pt,twocolumn,letterpaper]{article}
\PassOptionsToPackage{table}{xcolor}
\usepackage[pagenumbers]{cvpr}
\usepackage{microtype,ragged2e,needspace}
\usepackage{tabularx,array,multirow,float,placeins,balance}
\usepackage[most]{tcolorbox}
\usepackage{fontawesome5}
\definecolor{ReportBlue}{HTML}{013A89}
\definecolor{cvprblue}{HTML}{013A89}
\definecolor{DRRule}{HTML}{8FAFC7}
\definecolor{DRLight}{HTML}{F0F6FC}
\definecolor{OursRowBlue}{HTML}{F0F6FC}
\definecolor{OursBackground}{HTML}{F0F6FC}
\definecolor{SubHeaderTint}{HTML}{F4F6F9}
\definecolor{AbstractBg}{HTML}{F7F9FC}
\definecolor{AbstractBorder}{HTML}{D4E1EE}
\definecolor{findingbg}{HTML}{EFF7FF}
\definecolor{findingborder}{HTML}{6BAED6}
\definecolor{BaselineBackground}{HTML}{F2F2F2}
\definecolor{PerceptionBlue}{HTML}{244C70}
\definecolor{ProactiveTeal}{HTML}{247568}

\usepackage[breaklinks,colorlinks,allcolors=cvprblue]{hyperref}

\AtBeginDocument{%
  \crefname{figure}{Fig.}{Figs.}%
  \Crefname{figure}{Fig.}{Figs.}%
  \crefname{table}{Table}{Tables}%
  \Crefname{table}{Table}{Tables}%
  \crefname{equation}{Eq.}{Eqs.}%
  \Crefname{equation}{Eq.}{Eqs.}%
  \crefformat{section}{\S#2#1#3}%
  \Crefformat{section}{\S#2#1#3}%
  \crefformat{subsection}{\S#2#1#3}%
  \Crefformat{subsection}{\S#2#1#3}%
}

\providecommand{\vs}{\emph{vs.}\xspace}

\newcommand{\bestscore}[1]{\textbf{#1}}
\newcommand{\secondscore}[1]{\textcolor{black}{\underline{#1}}}

\arrayrulecolor{DRRule}
\hypersetup{
  pdftitle={StreamDLM: Streaming Video Understanding with Diffusion Large Language Model},
  pdfauthor={Zifeng Wang, Jiangtao Wu, Junkun Chen, Jiacong Fang, Jingyi Liu, Haojie Zheng, Kaiqi Liu, Peixuan Zhang, Si Li, Boxin Shi, Shuchen Weng},
  pdfsubject={Technical Report},
  pdfdisplaydoctitle=true
}

\newlength{\suppTableGap}
\newcommand{\suppSetupTableStyle}{\linespread{1}\fontsize{8}{9.6}\selectfont\renewcommand{\arraystretch}{1.07}}
\newcommand{\suppResultsTableStyle}{\linespread{1}\fontsize{9.15}{10.75}\selectfont\renewcommand{\arraystretch}{1.10}}
\newlength{\suppResultsGap}
\newlength{\suppPanelGap}

\usepackage{fancyhdr}
\newcommand{\ReportRunningHead}{StreamDLM: Streaming Video Understanding with Diffusion Large Language Model}

\fancypagestyle{streamdlmreport}{%
  \fancyhf{}%
  \fancyhead[L]{\raisebox{-5.2pt}[0pt][0pt]{\normalfont\rmfamily\fontsize{8.5}{10}\selectfont\color{ReportBlue}\ReportRunningHead}}%
  \fancyhead[R]{\raisebox{-5.2pt}[0pt][0pt]{\includegraphics[height=7.4pt]{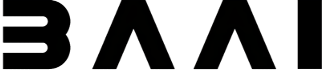}}}%
  \renewcommand{\headrulewidth}{.65pt}%
  \renewcommand{\headrule}{\hbox to\headwidth{\color{ReportBlue!85}\leaders\hrule height\headrulewidth\hfill}\vskip-\headrulewidth}%
  \fancyfoot[C]{\normalfont\rmfamily\fontsize{9}{10}\selectfont\color{black!75}\thepage}%
}

\fancypagestyle{streamdlmfirst}{%
  \fancyhf{}%
  \renewcommand{\headrulewidth}{0pt}%
  \fancyfoot[C]{\normalfont\rmfamily\fontsize{9}{10}\selectfont\color{black!68}\thepage}%
}

\begin{document}
\pagestyle{streamdlmreport}
\onecolumn
\thispagestyle{streamdlmfirst}

% ==================== Page 1 Masthead + Teaser + Soft Abstract Card ====================
\noindent\begin{minipage}{\textwidth}
\noindent
\begin{minipage}[c]{.27\linewidth}
  \includegraphics[height=27pt]{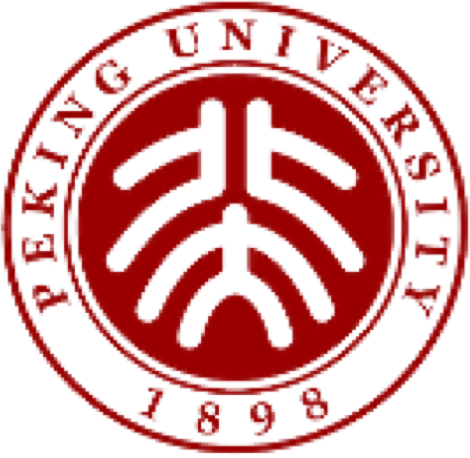}
\end{minipage}\hfill
\begin{minipage}[c]{.40\linewidth}\centering
  {\small\scshape\bfseries\color{ReportBlue} Technical Report}
\end{minipage}\hfill
\begin{minipage}[c]{.27\linewidth}\raggedleft
  \includegraphics[height=14.5pt]{sec/figures/baai-wordmark.png}
\end{minipage}
\par\vspace{4pt}
{\color{ReportBlue}\hrule height .85pt}
\vspace{6pt}
{\centering
  {\Large\bfseries StreamDLM: Streaming Video Understanding\\[2pt]
  with Diffusion Large Language Model\par}
\vspace{6pt}
  {\normalsize\bfseries
  Zifeng~Wang$^{1,2,*}$\quad Jiangtao~Wu$^{1,2,*}$\quad Junkun~Chen$^{1,2}$\quad Jiacong~Fang$^{1,2}$\quad Jingyi~Liu$^{1,2}$\\[0.2em]
  Haojie~Zheng$^{1,2}$\quad Kaiqi~Liu$^{1,2}$\quad Peixuan~Zhang$^{3,4}$\quad Si~Li$^{3,4}$\quad Boxin~Shi$^{1,\dagger}$\quad Shuchen~Weng$^{2,1,\dagger}$\par}
\vspace{0.38em}
  {\fontsize{8.5}{10.2}\selectfont\color{black!82}
  $^1$State Key Laboratory of Multimedia Information Processing, School of Computer Science, Peking University\\
  $^2$Beijing Academy of Artificial Intelligence\\
  $^3$School of Artificial Intelligence, Beijing University of Posts and Telecommunications\\
  $^4$Beijing Key Laboratory of Multimodal Data Intelligent Perception and Governance\par}
\vspace{0.28em}
  {\small\color{black!65} $^*$Equal contribution.\qquad $^\dagger$Corresponding authors.\par}
}%
\vspace{4pt}
\centering
\captionsetup{type=figure}
\includegraphics[width=.965\linewidth]{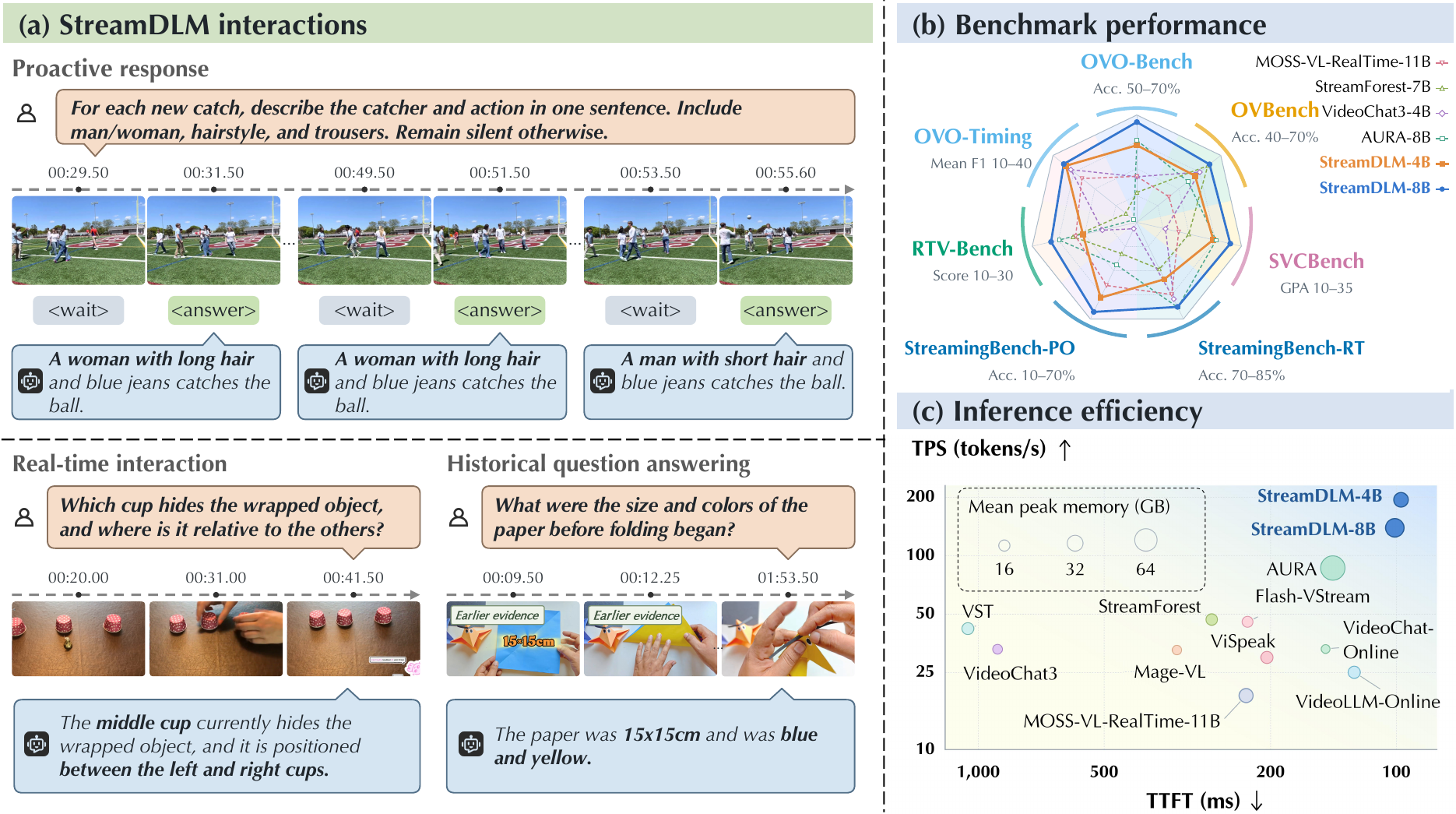}
\caption{\textbf{StreamDLM capabilities and performance.}
(a) Examples of proactive response, real-time interaction, and historical question answering.
(b) Performance on OVO-Bench~\cite{niu2025ovo}, OVBench~\cite{huang2025online}, SVCBench~\cite{liu2026svcbenchstreamingvideocounting}, StreamingBench-RT/PO~\cite{lin2026streamingbench}, and RTV-Bench~\cite{xun2026rtvbenchbenchmarkingmllmcontinuous}; OVO-Timing evaluates response timing (mean F1) on OVO-Bench following Em-Garde~\cite{zheng2026emgardeproposematchframeworkproactive}.
(c) Generation throughput (TPS) versus first-output latency (TTFT), where bubble area indicates mean peak GPU memory.
Across (b,c), StreamDLM outperforms streaming AR baselines~\cite{lu2026aura,zeng2026streamforest,li2026videochat3fullyopenvideo,wang2026mossvl,zhang2024flash,fu2025vispeak,yang2026magevlefficientcodecnativestreaming,huang2025online,chen2024videollm,guan2026video} in both accuracy and inference efficiency.}
\label{fig:teaser}
\par\vspace{5pt}
\begin{tcolorbox}[
  colback=AbstractBg,
  colframe=ReportBlue!55,
  arc=3.5pt,
  boxrule=0.6pt,
  left=10pt, right=10pt,
  top=5pt, bottom=6pt,
  before skip=0pt, after skip=0pt
]
  {\centering\normalsize\bfseries\color{ReportBlue} Abstract\par}
  \vspace{2.5pt}
  {\fontsize{9.2}{11.3}\selectfont\justifying\setlength{\parindent}{0pt}\noindent Streaming video understanding requires continuously perceiving ongoing events and interacting with users as frames arrive.
To achieve low-latency response generation, diffusion language models (DLMs) offer a compelling parallel alternative to sequential autoregressive (AR) decoding.
In this paper, we introduce StreamDLM, the first block-diffusion vision-language model for streaming video understanding.
To decide response timing, pre-answer gating separates response control from answer denoising via a gating step with class-balanced supervision.
To maintain context consistency, blockwise context management freezes the context within each block, refreshes it across blocks, and compresses and evicts older history.
To train efficiently over long-horizon streams, shared context packing packs multiple blocks over a de-duplicated shared KV prefix, reducing computational redundancy.
Across streaming benchmarks, StreamDLM achieves state-of-the-art performance in proactive response, real-time interaction, and historical question answering, with lower first-output latency and higher throughput than autoregressive streaming models.
\par}
\end{tcolorbox}
\end{minipage}%
\par
\twocolumn

\section{Introduction}
\label{sec:intro}

Streaming video understanding supports interactive applications such as wearable assistants and assistive robots~\cite{niu2025ovo,lin2026streamingbench,huang2025online}. These systems continuously receive video streams to perceive ongoing events, retain relevant historical context, and respond to queries grounded in accumulated visual evidence~\cite{wang2025proactivevideoqa,ma2026response}. Fig.~\ref{fig:teaser}~(a) illustrates these capabilities through proactive response to specified events, real-time question answering via continuous tracking, and historical question answering by recalling earlier attributes.

Most existing streaming systems rely on autoregressive (AR) vision-language models for response generation~\cite{chen2024videollm,lu2026aura,qian2025dispiderenablingvideollms}. However, the AR paradigm generates text token-by-token in a left-to-right manner, coupling decoding latency to sequence length and bottlenecking response completion in latency-sensitive interactions. In contrast, diffusion language models (DLMs) offer a compelling alternative through parallel token prediction and iterative refinement~\cite{NEURIPS2025_48b383b2,ye2508dream}. Specifically, block diffusion iteratively refines multi-token answer blocks via bidirectional attention, allowing unresolved positions to jointly attend to other tokens~\cite{arriola2025block}. By committing multiple high-confidence tokens per denoising step~\cite{wu2026fast}, DLMs reduce response completion latency without sacrificing generation quality.

To leverage these advantages, adapting DLMs to streaming video understanding requires resolving three key challenges:
\textit{(i) Response timing:}
The model continuously decides whether to respond. However, bidirectional attention exposes this timing decision to unmasked answer tokens during training, causing information leakage and shortcut learning.
\textit{(ii) Context consistency:}
Each response block requires iterative denoising within a fixed context budget, while the visual stream evolves continuously. Streaming context updates destabilize intra-block inference, whereas simply freezing the context ignores cross-block emerging dynamics.
\textit{(iii) Long-horizon training:}
Simulating causal history evolution over long sessions yields heavily overlapping context snapshots across interactions, where training each block independently incurs redundant computation.

In this paper, we propose \textbf{StreamDLM}, the first block-diffusion vision-language model for streaming video understanding, built on a pretrained diffusion VLM~\cite{chen2026bard} with decoupled observation and answer histories and three key designs:
\textit{(i) Pre-answer gating:}
To eliminate shortcut leakage from unmasked answer tokens, we decouple the response decision from answer denoising via a separate gating step conditioned on the current context snapshot. This design further enables independent loss normalization and class balancing for control supervision.
\textit{(ii) Blockwise context management:}
We maintain context consistency by freezing a context snapshot during intra-block denoising to provide a stable history. Across blocks, we bound the histories within a fixed budget via compression and eviction, and commit the previous block's states to the answer history, enabling the next block to receive a refreshed snapshot incorporating newly arrived inputs.
\textit{(iii) Shared context packing:}
For long-horizon training, we causally advance the observation and answer histories along the session timeline to construct context snapshots. We then pack the historical KV entries  into a de-duplicated shared prefix and train the model in a forward pass with segment-wise truncated backpropagation to reduce redundancy.

Experiments across five streaming video benchmarks demonstrate that StreamDLM outperforms previous methods in both perception and memory. Notably, StreamDLM achieves dominant performance in proactive response (65.60\% on StreamingBench-PO). On a single NVIDIA A100-SXM4-80GB GPU, StreamDLM-8B reduces first-output latency by 28.9\% and delivers a 1.61$\times$ generation speedup over the fastest AR baseline~\cite{lu2026aura}. Performance and efficiency gains are summarized in Fig.~\ref{fig:teaser}~(b) and~(c).

In summary, our contributions are threefold:
\begin{itemize}
    \item We propose StreamDLM, the first DLM-based streaming video understanding model,  supporting proactive response, real-time interaction, and historical question answering.
    \item We design pre-answer gating, blockwise context management, and shared context packing to resolve response timing, context consistency, and long-horizon training.
    \item Extensive experiments show that StreamDLM achieves superior perception, memory, and proactive response, with lower latency and higher throughput than AR baselines.
\end{itemize}

\section{Related Work}
\label{sec:related_work}

\subsection{Streaming Video Understanding}
Streaming video understanding pursues causal perception over continuously unfolding visual streams~\cite{yan2025streamformer,chen2024videollm}. For deciding response timing, prior methods attempt to decouple continuous perception from reasoning~\cite{qian2025dispiderenablingvideollms,yao2026joyaivlinteractionrealtimevisionlanguageinteraction,neupane2026foresight}, learn silence and response decisions within a unified backbone~\cite{lu2026aura,zeng2026onestreamer,fu2025vispeak,azad2026streamready}, or adopt interleaved interaction strategies that incorporate newly arriving frames between reasoning turns~\cite{wang2026mossvl,guan2026video}. To prevent unbounded memory growth over long horizons, existing works employ temporal token pruning, codec-guided selection~\cite{yao2025timechat,li2026videochat3fullyopenvideo,yang2026magevlefficientcodecnativestreaming}, structured memory banks~\cite{zhang2024flash,huang2025online,zeng2026streamforest}, or KV-cache compression~\cite{di2025rekv,xu2026streamingvlm,kim2025infinipotv,yang2025streammem}. However, existing streaming models rely on autoregressive decoders, where token-by-token generation bottlenecks real-time interaction.
Our StreamDLM introduces block diffusion for parallel token prediction, significantly reducing latency and improving throughput.

\subsection{Diffusion Vision-Language Models}
Discrete diffusion language models generate text through iterative unmasking and refinement~\cite{NEURIPS2025_48b383b2,ye2508dream}, while block diffusion combines intra-block parallel prediction with cross-block causal dependencies~\cite{arriola2025block}. Recent works construct diffusion VLMs either by extending pretrained diffusion language models via multimodal training~\cite{you2026llada,li2025lavida,cheng2026sdarvl,ye2025dreamvl,yu2025dimple} or by adapting autoregressive backbones into diffusion paradigms~\cite{zeng2025diffusionvl,wu2026fastdvlm,chen2026bard}. For video understanding, VidLaDA accelerates decoding over offline video inputs~\cite{he2026vidlada}, whereas STRIDE only denoises activation decisions to trigger a frozen downstream model~\cite{kim2026stride}.
To bring DLM advantages to end-to-end streaming video understanding, our StreamDLM builds upon a pretrained block-diffusion VLM~\cite{chen2026bard} and integrates pre-answer gating with blockwise context management to unify response timing and iterative generation.

\subsection{Efficient Diffusion Decoding}
Efficient diffusion decoding aims to reduce iterative computation and coordinate parallel token updates. Recent methods accelerate inference via KV-cache reuse and hierarchical block caching~\cite{wu2026fast,wu2026fastdllmv2}, confidence-guided parallel commitment and early termination~\cite{wu2026fast,li2026prophet}, or adaptive sampling and bidirectional token verification~\cite{wei2026slowfast,hong2026wino}. Other works further improve generation quality by aggregating predictions across denoising steps~\cite{wang2026timefeature} or optimizing token decoding orders~\cite{ni2026flexibility}.
However, these acceleration strategies assume a static conditioning context, which breaks down when visual inputs evolve continuously during streaming inference.
In contrast, StreamDLM adopts blockwise context management with frozen intra-block snapshots to enable stable iterative diffusion decoding over evolving streams.

\begin{figure*}[!t]
    \centering
    \includegraphics[width=\linewidth]{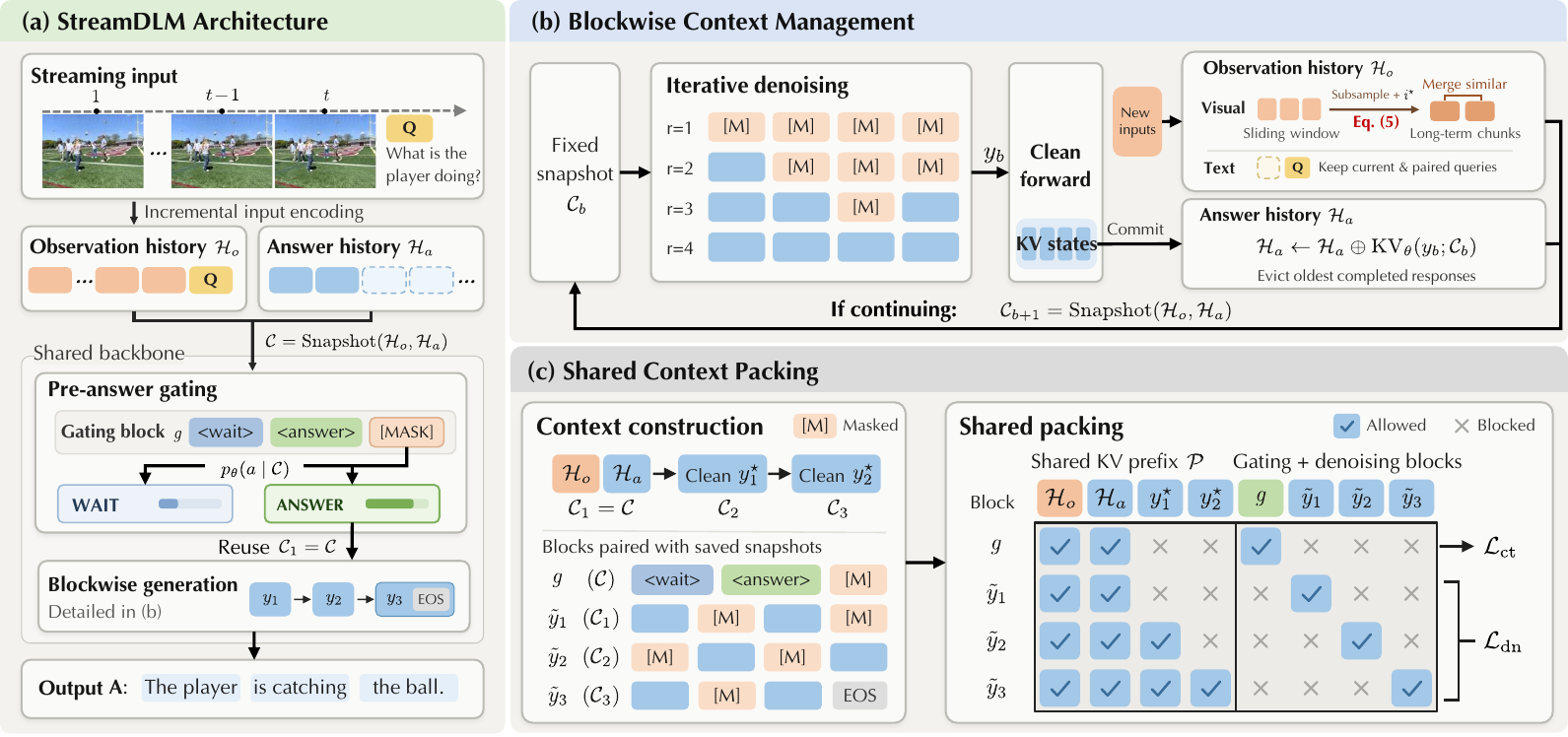}
\caption{
\textbf{Overview of StreamDLM.}
\textbf{(a) StreamDLM architecture and pre-answer gating:} The model maintains decoupled observation ($\mathcal{H}_o$) and answer ($\mathcal{H}_a$) histories and shares a backbone across pre-answer gating and blockwise generation. A gating step predicts \texttt{WAIT} to defer answering or \texttt{ANSWER} to initiate generation, with the first answer block reusing the gating snapshot ($\mathcal{C}_1=\mathcal{C}$).
\textbf{(b) Blockwise context management:} The snapshot $\mathcal{C}_b$ stays fixed during intra-block iterative denoising and is refreshed across block boundaries after clean KV commitment, visual compression, and oldest-response eviction within a fixed budget.
\textbf{(c) Shared context packing:} Training constructs causal snapshots as reference blocks update the history (left) and packs the paired gating and noisy blocks over a de-duplicated shared KV prefix $\mathcal{P}$ in one forward pass (right), using a structured attention mask to supervise control ($\mathcal{L}_{\mathrm{ct}}$) and denoising ($\mathcal{L}_{\mathrm{dn}}$). 
}
    \label{fig:pipeline}
    \vspace{-1em}
\end{figure*}

\section{Methodology}
\label{sec:method}

In this section, we present \textbf{StreamDLM}, the first block-diffusion vision-language model for streaming video understanding, as illustrated in \Cref{fig:pipeline}.

\subsection{StreamDLM Architecture}
\label{sec:method_prelim}

StreamDLM builds on a pretrained diffusion VLM~\cite{chen2026bard}, in which a visual encoder and a visual merger map video frames into visual tokens, and a block-diffusion language model generates responses conditioned on visual and textual tokens.

\noindent\textbf{Blockwise generation.}
Given a multimodal context $\mathbf{x}$, block diffusion~\cite{arriola2025block} generates a response $\mathbf{y}=(\mathbf{y}_1,\ldots,\mathbf{y}_K)$ autoregressively across blocks while predicting tokens in parallel within each block:
\begin{equation}
    p_\theta(\mathbf{y}\mid\mathbf{x})
    = \prod_{b=1}^{K}
           p_\theta(\mathbf{y}_b\mid\mathbf{x},\mathbf{y}_{<b}),
    \label{eq:method_block_factorization}
\end{equation}
where $\theta$ denotes the model parameters, $\mathbf{y}_{<b}=(\mathbf{y}_1,\ldots,\mathbf{y}_{b-1})$ denotes the preceding blocks in the current answer, and each block $\mathbf{y}_b$ contains $B$ token positions.
Each block $\mathbf{y}_b$ is initialized with $B$ \texttt{[MASK]} tokens and generated through iterative denoising.
At each denoising step, the model predicts token distributions for all remaining masked positions in parallel, applying bidirectional attention within $\mathbf{y}_b$ and causal attention to $\mathbf{x}$ and $\mathbf{y}_{<b}$.
The positions with the highest prediction confidence are then unmasked, while previously unmasked tokens remain fixed in subsequent steps.
This process repeats until all positions in $\mathbf{y}_b$ are resolved, after which the completed block serves as context for generating $\mathbf{y}_{b+1}$.
Generation terminates once a completed block contains an end-of-sequence (\texttt{[EOS]}) token or upon reaching the maximum generation length. %

\noindent\textbf{Decoupled histories.}
Eq.~\eqref{eq:method_block_factorization} assumes that the input $\mathbf{x}$ is fully available before generation begins.
In streaming interaction, however, video frames and user queries arrive continuously, and earlier responses also become part of the context for subsequent interactions.
As shown in \cref{fig:pipeline}~(a), StreamDLM maintains two persistent key--value (KV) caches throughout a session: an observation history $\mathcal{H}_o$ and an answer history $\mathcal{H}_a$.
The observation history $\mathcal{H}_o$ stores the KV states of incoming video frames and user queries.
Each newly arrived frame or query is encoded using only $\mathcal{H}_o$ as causal context, and its resulting KV states are further appended to $\mathcal{H}_o$.
Meanwhile, the answer history $\mathcal{H}_a$ stores the KV states of committed answer blocks across the session.
Decoupling the two histories keeps visual and query encoding independent of model-generated answers, allowing each cache to be updated and bounded separately.

\noindent\textbf{Context snapshot.}
To condition block-diffusion generation on these evolving histories, we define a context snapshot as a frozen view of both KV caches at a given timestamp:
\begin{equation}
    \mathcal{C}=\operatorname{Snapshot}(\mathcal{H}_o,\mathcal{H}_a).
    \label{eq:method_snapshot}
\end{equation}
Once captured, a snapshot is unaffected by subsequent updates to the histories.
Denoting the snapshot captured for answer block $\mathbf{y}_b$ by $\mathcal{C}_b$, the blockwise generation in Eq.~\eqref{eq:method_block_factorization} naturally adapts to the streaming setting by replacing the static context $(\mathbf{x},\mathbf{y}_{<b})$ with $\mathcal{C}_b$, yielding $p_\theta(\mathbf{y}_b\mid\mathcal{C}_b)$.
Subsequently, we introduce both pre-answer gating (\cref{sec:method_control}) and blockwise context management (\cref{sec:method_architecture}), which operate on snapshots of the retained histories.

\subsection{Pre-Answer Gating}
\label{sec:method_control}

A streaming model needs to continuously decide whether to respond as the video stream evolves.
An approach is to place a control token inside the answer block and predict it jointly with the answer.
However, under random masking during training, bidirectional attention within the block allows this control token to attend to unmasked answer tokens, causing information leakage and shortcut learning.

\noindent\textbf{Gating step.}
To eliminate this leakage, pre-answer gating decouples the response decision from answer generation by moving it to a separate gating step before answer denoising begins.
At each gating step, given the current context snapshot $\mathcal{C}$ (Eq.~\eqref{eq:method_snapshot}), the model performs a forward pass on a gating block $\mathbf{g}=[\texttt{<wait>}, \texttt{<answer>}, \texttt{[MASK]}]$ using the shared backbone, where \texttt{<wait>} and \texttt{<answer>} are special control tokens.
Let $\mathcal{A}=\{\texttt{WAIT},\texttt{ANSWER}\}$ denote the candidate action set, and let $z_{\texttt{WAIT}}$ and $z_{\texttt{ANSWER}}$ denote the logits of \texttt{<wait>} and \texttt{<answer>} predicted at the \texttt{[MASK]} position.
The action probability is given by
\begin{equation}
    p_\theta(a \mid \mathcal{C})
    =
    \frac{\exp(z_a)}{\sum_{a'\in\mathcal{A}}\exp(z_{a'})},
    \quad a\in\mathcal{A},
    \label{eq:method_gating}
\end{equation}
and the model selects the action with the higher probability.
Because this decision is conditioned only on $\mathcal{C}$ and $\mathbf{g}$, no tokens of the current answer are exposed during training or inference.
Finally, $\mathbf{g}$ is discarded after the gating step without occupying KV cache in either history.

\noindent\textbf{Decision execution.}
The streaming loop then branches according to the selected action, as illustrated in \cref{fig:pipeline}(a).
If \texttt{WAIT} is selected, newly arriving video frames and user queries continue to be appended to $\mathcal{H}_o$, and a refreshed snapshot is evaluated at the next gating step.
If \texttt{ANSWER} is selected, the current snapshot is directly reused as $\mathcal{C}_1=\mathcal{C}$ to initiate denoising for the first answer block $\mathbf{y}_1$ (\cref{sec:method_architecture}).
Beyond inference control, separating this dedicated gating step from answer tokens also enables explicit loss normalization and class balancing during training (\cref{sec:method_training_details}).

\subsection{Blockwise Context Management}
\label{sec:method_architecture}

Each answer block requires multiple denoising steps, during which new video frames and user queries continue to arrive.
If these streaming context updates were exposed during iterative denoising, the conditioning context would shift mid-block and destabilize intra-block generation.
Conversely, freezing the context across the entire response would ignore emerging visual dynamics.
As shown in \cref{fig:pipeline}(b), blockwise context management freezes the snapshot $\mathcal{C}_b$ within each block and refreshes it at block boundaries after clean KV commitment, bounding histories within a fixed budget.

\noindent \textbf{Snapshot refreshing.}
Starting from the initial snapshot $\mathcal{C}_1=\mathcal{C}$ provided by pre-answer gating (\cref{sec:method_control}), we hold $\mathcal{C}_b$ fixed across all denoising steps of block $\mathbf{y}_b$.
Meanwhile, newly arriving frames and user queries are still encoded and appended to $\mathcal{H}_o$, without perturbing the ongoing denoising of $\mathbf{y}_b$.
Once all positions in $\mathbf{y}_b$ are unmasked, the block is truncated after the first \texttt{[EOS]} token (if present) to form the finalized block.
Since the KV states from the final denoising step were computed while some positions were still masked, they cannot be directly reused.
We therefore run a forward pass over $\mathbf{y}_b$ conditioned on the same snapshot $\mathcal{C}_b$ and commit its KV states to the answer history:
\begin{equation}
    \mathcal{H}_{a}
    \leftarrow
    \mathcal{H}_{a} \oplus
    \operatorname{KV}_{\theta}(\mathbf{y}_b; \mathcal{C}_b),
    \label{eq:method_commit}
\end{equation}
where $\operatorname{KV}_{\theta}(\mathbf{y}_b; \mathcal{C}_b)$ denotes the KV states of $\mathbf{y}_b$ computed under $\mathcal{C}_b$, and $\oplus$ denotes concatenation along the sequence dimension.
The committed block is then emitted to the user.
If generation continues, a refreshed snapshot $\mathcal{C}_{b+1}$ is captured from the updated histories via Eq.~\eqref{eq:method_snapshot}, incorporating both the inputs arriving during the denoising of $\mathbf{y}_b$ and all retained answer blocks.

\noindent \textbf{Visual history.}
To support unbounded streams within a fixed context budget, we bound the combined token count of $(\mathcal{H}_o, \mathcal{H}_a)$ before taking each snapshot, reserving $B$ tokens for the upcoming answer block.
For visual KV states in $\mathcal{H}_o$, recent frames are kept in full within a sliding window.
As an older video segment exits this window, we compress its visual KV states into a compact long-term chunk.
Specifically, we first uniformly subsample frames within the segment, and then partition each retained frame's visual tokens into contiguous intervals of roughly equal size.
Within each interval $I$, we select the single token position that deviates most from the local interval mean:
\begin{equation}
    i^\star =
    \arg\max_{i\in I}
    \left\|\mathbf{v}_i-\bar{\mathbf{v}}_I\right\|_2^2,
    \label{eq:method_visual_selection}
\end{equation}
where $\mathbf{v}_i$ is the $\ell_2$-normalized concatenation of last-layer value vectors across attention heads at position $i$, and $\bar{\mathbf{v}}_I$ is their mean over $I$.
Selecting the most distinctive token relative to the local mean suppresses homogeneous redundancy while preserving spatial coverage across the frame.
The selected positions are retained across all KV layers with their original positional encodings.
When the long-term visual budget is exceeded, we compute the cosine similarity between adjacent chunks using their mean last-layer value vectors, and merge the most similar pair into a chunk spanning a longer duration by re-applying the compression approach.

\noindent\textbf{Textual history.}
For textual KV states, we always retain the current user query, the queries corresponding to non-evicted historical answers in $\mathcal{H}_o$, and the ongoing response in $\mathcal{H}_a$.
To stay within the respective token budgets, older standalone input text is dropped first, and the oldest completed answer turns are evicted when necessary.

\subsection{Shared Context Packing}
\label{sec:method_training}

Long-horizon training requires gating and denoising blocks to use contexts constructed under the exact same rules as inference.
However, since causal interactions across a video stream share heavily overlapping histories, evaluating each block independently incurs redundant computation over repeated KV states.
As illustrated in \Cref{fig:pipeline}~(c), we therefore design shared context packing, which simulates session-level history evolution and packs multiple interactions over a shared historical prefix into a single forward pass.

\noindent \textbf{Context construction.}
Each training session pairs a stream of video frames and user queries with reference answers, where each answer is partitioned into $K$ blocks $\mathbf{y}^\star=(\mathbf{y}_1^\star,\ldots,\mathbf{y}_K^\star)$ terminated with \texttt{[EOS]}.
For tasks requiring response timing, control annotations $a\in\mathcal{A}$ are further provided.
To simulate the inference setting, we causally advance the observation and answer histories $(\mathcal{H}_o,\mathcal{H}_{a})$ along the session timeline.
At each supervised control decision, the current snapshot $\mathcal{C}$ is paired with the gating block $\mathbf{g}$ to supervise pre-answer gating without modifying either history.
When a response is triggered, the current $\mathcal{C}$ is bound to its first block as $\mathcal{C}_1$, while subsequent block snapshots follow the history updates and compression rules in \cref{sec:method_architecture}.
Specifically, complete reference blocks advance the answer history (\cref{fig:pipeline}~(c), left):
\begin{equation}
    \mathcal{H}_a\leftarrow\mathcal{H}_a\oplus\operatorname{KV}_\theta(\mathbf{y}_b^\star;\mathcal{C}_b),\qquad b=1,\ldots,K.
    \label{eq:training_snapshot_blocks}
\end{equation}
Each block $\mathbf{y}_b^\star$ is then masked at a sampled ratio $\rho_b\in(0,1]$, and the resulting noisy block $\tilde{\mathbf{y}}_b$ is paired with $\mathcal{C}_b$ (captured prior to committing $\mathbf{y}_b^\star$), ensuring that no future tokens from $\mathbf{y}_{\ge b}^\star$ leak into its conditioning context.

\noindent \textbf{Shared packing.}
To avoid redundant context processing, we collect all historical KV entries across the session into a de-duplicated shared KV prefix $\mathcal{P}$. All gating and denoising blocks are then concatenated after $\mathcal{P}$ to form a packed sequence (\cref{fig:pipeline}~(c), right).
For each packed block covering token positions $U$, let $\mathcal{I}_U$ denote the subset of positions in $\mathcal{P}$ belonging to its saved snapshot ($\mathcal{C}$ for a gating block $\mathbf{g}$, or $\mathcal{C}_b$ for a denoising block $\tilde{\mathbf{y}}_b$).
We construct an attention mask over all packed positions $j$:
\begin{equation}
    M_{ij}=\begin{cases}
        0, & \text{if } j\in\mathcal{I}_U\cup U,\\
        -\infty, & \text{otherwise},
    \end{cases}\qquad i\in U,
    \label{eq:training_packed_mask}
\end{equation}
which enforces each block's exact historical visibility and block-diagonal bidirectional attention within each block, enabling all gating and denoising predictions in the session to be evaluated in a forward pass.
Finally, to bound activation memory over long streams, we apply segment-wise truncated backpropagation\footnote{Implementation details are provided in \cref{sec:supp_method}.}, partitioning the session into segments and detaching historical KV states at segment boundaries while accumulating gradients across segments.

\begin{table*}[!t]
\centering
\caption{
Performance comparison of offline and streaming video models across video perception, memory, and proactive response tasks.
Higher scores indicate better performance.
\textbf{Boldface} and \underline{underlining} denote the best and second-best results among streaming models, respectively.
}
\label{tab:main_results}

\begingroup
\small
\setlength{\tabcolsep}{2.5pt}
\renewcommand{\arraystretch}{1.15}

\newcommand{\MainTableRows}{%

\toprule
\multirow{4}{*}[-0.6ex]{\textbf{Model}} &
\multirow{4}{*}[-0.6ex]{\textbf{Params}} &
\multicolumn{7}{c}{%
  \textbf{Perception \& Memory}%
} &
\multicolumn{2}{c}{%
  \textbf{Proactive Response}%
} \\
\cmidrule(lr){3-9}
\cmidrule(lr){10-11}
& &
\shortstack{OVO-\\Bench} &
\shortstack{Streaming\\Bench} &
\raisebox{0.5\normalbaselineskip}{OVBench} &
\multicolumn{3}{c}{\raisebox{0.5\normalbaselineskip}{SVCBench}} &
\shortstack{RTV-\\Bench} &
\shortstack{Streaming\\Bench} &
\shortstack{OVO-\\Timing} \\
\cmidrule(lr){3-3}
\cmidrule(lr){4-4}
\cmidrule(lr){5-5}
\cmidrule(lr){6-8}
\cmidrule(lr){9-9}
\cmidrule(lr){10-10}
\cmidrule(lr){11-11}
& & Overall & Real-Time & Avg.
& GPA & MoC & UDA
& Score & PO & Mean F1 \\
\midrule

\rowcolor{SubHeaderTint}
\multicolumn{11}{l}{\textit{Offline Video Models}} \\
LLaVA-OneVision~\cite{li2024llava}
& 7B
& 52.90 & 71.10 & 49.50
& 16.21 & 30.15 & 9.82
& 4.40 & 29.55 & 10.88 \\

Qwen2.5-VL~\cite{bai2025qwen25vl}
& 7B
& 50.50 & 73.68 & 50.61
& 19.10 & 68.20 & 40.80
& 7.13 & 23.20 & 7.98 \\

InternVL3.5~\cite{internvl2025internvl35}
& 8B
& 53.78 & 77.96 & 50.96
& 24.20 & 81.60 & 55.60
& 20.26 & 42.80 & 20.52 \\

MiniCPM-o 4.5~\cite{cui2026minicpm}
& 9B
& 59.70 & 78.20 & 49.30
& 28.70 & 78.92 & 51.22
& 20.22 & 47.60 & 21.30 \\

\midrule
\rowcolor{SubHeaderTint}
\multicolumn{11}{l}{\textit{Streaming Video Models}} \\

Flash-VStream~\cite{zhang2024flash}
& 7B
& 33.20 & 23.20 & 31.20
& 23.40 & 78.50 & 34.70
& 10.79 & 1.96 & 4.77 \\

VideoLLM-Online~\cite{chen2024videollm}
& 8B
& 21.12 & 36.00 & 9.60
& 10.94 & 70.31 & 22.81
& 4.83 & 3.92 & 6.93 \\

VideoChat-Online~\cite{huang2025online}
& 4B
& 40.40 & 58.81 & 54.90
& 18.84 & 82.60 & 33.36
& 17.58 & 1.20 & 24.30 \\

Dispider~\cite{qian2025dispiderenablingvideollms}
& 7B
& 41.80 & 67.60 & 43.85
& 7.70 & 34.80 & 15.50
& 14.25 & 2.00 & 5.98 \\

ViSpeak~\cite{fu2025vispeak}
& 7B
& 59.40 & 70.40 & 45.24
& 22.37 & 66.37 & 42.64
& 14.36 & 44.40 & 31.87 \\

TimeChat-Online~\cite{yao2025timechat}
& 7B
& 45.70 & 75.40 & 45.48
& 22.55 & 80.15 & 35.59
& 14.60 & 8.40 & 5.51 \\

StreamForest~\cite{zeng2026streamforest}
& 7B
& 55.57 & 77.26 & \secondscore{65.37}
& 22.41 & 83.27 & 37.36
& 22.21 & 29.50 & 13.95 \\

StreamingVLM~\cite{xu2026streamingvlm}
& 7B
& 50.93 & 73.20 & 47.80
& 19.10 & 68.10 & \bestscore{50.30}
& 14.40 & 1.20 & 11.14 \\

VST~\cite{guan2026video}
& 7B
& 59.30 & 79.50 & 49.18
& 20.01 & 66.17 & 43.96
& 17.90 & 4.00 & 19.15 \\

AURA~\cite{lu2026aura}
& 8B
& \secondscore{65.30} & \bestscore{83.20} & 58.30
& 28.91 & 74.43 & 39.86
& \secondscore{24.84} & 36.40 & 11.10 \\

JoyAI-VL-Interaction~\cite{yao2026joyaivlinteractionrealtimevisionlanguageinteraction}
& 8B
& 59.10 & 82.70 & 62.30
& \secondscore{30.82} & 82.66 & \secondscore{49.80}
& 20.57 & 35.20 & 20.00 \\

VideoChat3~\cite{li2026videochat3fullyopenvideo}
& 4B
& 58.50 & 81.90 & 62.50
& 16.84 & 62.29 & 34.93
& 16.60 & 14.00 & 33.60 \\

Mage-VL~\cite{yang2026magevlefficientcodecnativestreaming}
& 4B
& 58.50 & 82.10 & 57.50
& 26.78 & 56.31 & 34.33
& 20.64 & 6.80 & 21.70 \\

Foresight~\cite{neupane2026foresight}
& 8B
& 62.16 & 79.30 & 52.88
& 24.93 & \bestscore{84.55} & 47.28
& 17.50 & 41.60 & 20.60 \\

MOSS-VL-Realtime~\cite{wang2026mossvl}
& 11B
& 58.56 & 81.20 & 51.46
& 19.99 & 77.95 & 37.40
& 21.70 & 49.20 & 29.64 \\

\midrule
\rowcolor{SubHeaderTint}
\multicolumn{11}{l}{\textit{Base VLM Backbones}} \\

Bard-VL-4B~\cite{chen2026bard}
& 4B
& 47.99 & 72.04 & 51.82
& 21.56 & 82.72 & 41.75
& 6.62 & 37.20 & 2.41 \\

Bard-VL-8B~\cite{chen2026bard}
& 8B
& 48.55 & 73.16 & 52.34
& 21.65 & 65.08 & 40.87
& 7.68 & 37.20 & 0.40 \\

Qwen3-VL~\cite{bai2025qwen3}
& 8B
& 46.80 & 74.10 & 52.00
& 31.00 & 84.30 & 55.10
& 17.83 & 52.80 & 14.27 \\

\midrule

\rowcolor{OursRowBlue}
\textbf{StreamDLM-4B}
& 4B
& 64.36 & 78.84 & 60.83
& 28.25 & 71.28 & 37.66
& 20.30 & \secondscore{56.80} & \secondscore{35.35} \\

\rowcolor{OursRowBlue}
\textbf{StreamDLM-8B}
& 8B
& \bestscore{68.68} & \secondscore{83.12} & \bestscore{65.99}
& \bestscore{32.33} & \secondscore{83.53} & 44.59
& \bestscore{26.41} & \bestscore{65.60} & \bestscore{36.14} \\

\bottomrule

}
\newsavebox{\MainTableMeasure}
\sbox{\MainTableMeasure}{\begin{tabular}{lcccccccccc}\MainTableRows\end{tabular}}
\newlength{\MainTableExtraSpace}
\setlength{\MainTableExtraSpace}{\dimexpr\textwidth-\wd\MainTableMeasure\relax}
\divide\MainTableExtraSpace by 22
\addtolength{\tabcolsep}{\MainTableExtraSpace}
\begin{tabular}{lcccccccccc}
\MainTableRows
\end{tabular}
\endgroup
\end{table*}

\subsection{Training Objective}
\label{sec:method_training_details}
Since pre-answer gating separates response decisions from answer generation, we normalize the control and denoising objectives independently.

\noindent\textbf{Control loss.}
To address the class imbalance between frequent \texttt{WAIT} and sparse \texttt{ANSWER} decisions, we formulate $\mathcal{L}_{\mathrm{ct}}$ as a class-balanced cross-entropy loss:
\begin{equation}
    \mathcal{L}_{\mathrm{ct}}
    = -\frac{1}{|\mathcal{A}_s|}
    \sum_{a\in\mathcal{A}_s}
    \frac{1}{|\mathcal{D}_a|}
    \sum_{\mathcal{C}\in\mathcal{D}_a}
    \log p_\theta(a\mid \mathcal{C}),
    \label{eq:training_session_losses}
\end{equation}
where $\mathcal{A}_s\subseteq\mathcal{A}$ denotes the action classes present in the session, and $\mathcal{D}_a$ is the multiset of context snapshots saved at control decisions labeled with action $a$, counting one entry per decision even when snapshots are identical ($\mathcal{L}_{\mathrm{ct}}$ is omitted for sessions without control annotations).

\noindent\textbf{Denoising loss.}
For answer generation, we adopt the inverse-rate weighted masked cross-entropy loss~\cite{arriola2025block,chen2026bard} over each noisy block $\tilde{\mathbf{y}}_b$:
\begingroup
\small
\setlength{\abovedisplayskip}{5pt}
\setlength{\abovedisplayshortskip}{5pt}
\setlength{\belowdisplayskip}{5pt}
\setlength{\belowdisplayshortskip}{5pt}
\begin{equation}
\mathcal{L}_{\mathrm{dn}}
\!=\! -\frac{1}{N_{\mathrm{ans}}}
\sum_{\mathbf{y}^\star}
\frac{1}{|\mathbf{y}^\star|}
\sum_{b=1}^{K}
\frac{1}{\rho_b}
\sum_{i\in\mathcal{M}_b}
\!\log p_\theta(y_{b,i}^\star \!\mid\! \tilde{\mathbf{y}}_b, \mathcal{C}_b),
\label{eq:training_denoising_loss}
\end{equation}
\endgroup
where $\mathcal{M}_b$ denotes the set of non-padding masked token positions in block $\tilde{\mathbf{y}}_b$, $\vert{}\mathbf{y}^\star\vert{}$ is the valid answer length including \texttt{[EOS]}, and $N_{\mathrm{ans}}$ is the number of reference answers with content supervision in the session.

\noindent\textbf{Total loss.}
We calculate both losses from the packed forward pass and jointly optimize them as $\mathcal{L}=(\mathcal{L}_{\mathrm{ct}}+\mathcal{L}_{\mathrm{dn}})/2$ when both objectives are supervised. Sessions with only one supervised objective optimize the corresponding loss alone.

\section{Experiments}

\subsection{Experimental Setup}
\label{sec:exp_setup}

\noindent\textbf{Training data.}
Our training corpus comprises 598{,}400 streaming sessions constructed from public video datasets and instruction corpora, primarily including LLaVA-Video~\cite{zhang2024llava}, LaSOT~\cite{fan2019lasot}, GOT-10k~\cite{huang2019got}, Shot2Story~\cite{han2025shot2story}, and STAR~\cite{wu2024star}.
In total, these sessions provide approximately 10.92M control decisions and 1.90M reference answers.
To support long-horizon context management and multi-turn interaction, 23.81\% of the sessions exceed 80\,s in duration, and 26.06\% contain multiple reference answers.
Control annotations are available in 36.66\% of the sessions, consisting of 9.52M \texttt{WAIT} and 1.40M \texttt{ANSWER} labels.

\noindent\textbf{Implementation details.}
We initialize StreamDLM-4B and StreamDLM-8B from pretrained Bard-VL~\cite{chen2026bard} backbones and train them on 32 NVIDIA H100 GPUs with a global batch size of 32 sessions.
Both models sample video streams at 2\,FPS with a per-frame budget of 131{,}072 pixels, operating under a fixed context budget of 32{,}768 tokens and an answer block size of $B=4$.
During training, we keep the visual encoder frozen while updating the visual merger and the embeddings of the special control tokens (\texttt{<wait>} and \texttt{<answer>}).
For the language backbone, we apply full-parameter fine-tuning for StreamDLM-4B and rank-128 LoRA adapters for StreamDLM-8B.

\noindent\textbf{Benchmarks.}
We evaluate StreamDLM across 5 streaming benchmarks covering 3 core capabilities: proactive response, real-time interaction, and historical question answering.
Specifically, OVO-Bench~\cite{niu2025ovo} and OVBench~\cite{huang2025online} assess current-event perception, historical recall, and forward-looking reasoning, while the real-time (RT) split of StreamingBench~\cite{lin2026streamingbench} evaluates online perception and reasoning over unfolding streams.
SVCBench~\cite{liu2026svcbenchstreamingvideocounting} and RTV-Bench~\cite{xun2026rtvbenchbenchmarkingmllmcontinuous} test the ability to track evolving scene states through continuous counting and time-sensitive question answering, respectively.
Finally, the Proactive Output (PO) split of StreamingBench and the OVO-Timing protocol~\cite{zheng2026emgardeproposematchframeworkproactive} on OVO-Bench evaluate proactive response quality and temporal trigger accuracy.

\noindent\textbf{Baselines.}
We compare StreamDLM against 4 offline and 15 streaming video understanding models spanning diverse architectures, memory mechanisms, and response control strategies.
To evaluate the benefits of our streaming block-diffusion framework, we also report results for the base Bard-VL~\cite{chen2026bard} and Qwen3-VL~\cite{bai2025qwen3} models.
Additional details on dataset construction, hyperparameters, and evaluation protocols are provided in \cref{sec:supp_training}.

\subsection{Main Results}
\label{sec:main_results}

\noindent\textbf{Perception and memory.}
As shown in \cref{tab:main_results}, StreamDLM achieves state-of-the-art performance across online perception, historical recall, and continuous tracking benchmarks.

Among streaming models, StreamDLM-8B ranks first on OVO-Bench (68.68\%), OVBench (65.99\%), SVCBench GPA (32.33), and RTV-Bench (26.41), while matching the top AR baseline AURA~\cite{lu2026aura} on StreamingBench-RT (83.12\% vs.\ 83.20\%).

\noindent\textbf{Proactive response.}
StreamDLM-8B establishes a dominant lead on proactive tasks that require deciding both when and what to respond.
Specifically, StreamDLM-8B reaches 65.60\% accuracy on StreamingBench-PO, outperforming the strongest 11B streaming competitor MOSS-VL-Realtime~\cite{wang2026mossvl} (49.20\%) by 16.40\%, and achieves the highest mean F1 of 36.14 on OVO-Timing.

\noindent\textbf{Comparison with base backbones.}
StreamDLM-8B improves over its pretrained Bard-VL-8B~\cite{chen2026bard} backbone by 20.13\% on OVO-Bench and 28.40\% on StreamingBench-PO, and outperforms Qwen3-VL-8B~\cite{bai2025qwen3}, the base model from which Bard-VL-8B is trained. These gains reflect the effectiveness of our streaming adaptation and the inherent strengths of block-diffusion generation.

\noindent\textbf{Model scaling and parameter efficiency.}
Our lightweight StreamDLM-4B (64.36\% on OVO-Bench and 60.83\% on OVBench) already surpasses most 7B--11B streaming AR baselines across perception and memory tasks, and ranks second only to StreamDLM-8B on both StreamingBench-PO (56.80\%) and OVO-Timing (35.35).
Scaling the backbone to 8B further boosts accuracy across all benchmarks, demonstrating parameter efficiency and scaling gains.

\begin{table*}[!t]
    \centering
    \begin{minipage}[t]{0.49\textwidth}
        \vspace{0pt}
        \centering
        \caption{Effect of context management. Str.\ denotes StreamingBench. OVBench reports question-level accuracy within each category.}
        \label{tab:temporal_window}

        \small
        \setlength{\tabcolsep}{1pt}
        \renewcommand{\arraystretch}{1.18}
        \begin{tabularx}{\linewidth}{
            @{}l
            *{6}{>{\centering\arraybackslash}X}
            @{}
        }
            \toprule
            & \multicolumn{2}{c}{OVO-Bench}
            & \multicolumn{3}{c}{OVBench}
            & Str. \\
            \cmidrule(lr){2-3}
            \cmidrule(lr){4-6}
            \cmidrule(lr){7-7}
            Configuration & RVP & FAR & SP & FP & PM & PO \\
            \midrule
            4\,s window
            & \textbf{75.00} & 47.70
            & \textbf{66.72} & 54.81 & 62.58
            & 58.40 \\
            16\,s window
            & 72.31 & 53.85
            & 62.21 & 56.01 & 64.16
            & 65.20 \\
            80\,s window
            & 70.08 & 59.23
            & 61.28 & 57.39 & 68.42
            & \textbf{65.60} \\
            \rowcolor{OursRowBlue}
            \textbf{Ours}
            & 71.10 & \textbf{60.71}
            & 60.65 & \textbf{58.23} & \textbf{69.06}
            & \textbf{65.60} \\
            \bottomrule
        \end{tabularx}
    \end{minipage}
    \hfill
    \begin{minipage}[t]{0.49\textwidth}
        \vspace{0pt}
        \centering
        \caption{Performance of different block sizes $B$. OVBench reports the macro-average over 16 subtasks. }
        \label{tab:block_size}

        \small
        \setlength{\tabcolsep}{1pt}
        \renewcommand{\arraystretch}{1.18}
        \begin{tabularx}{\linewidth}{
            @{}l
            *{5}{>{\centering\arraybackslash}X} >{\centering\arraybackslash}p{.16\linewidth}
            @{}
        }
            \toprule
            & \multicolumn{3}{c}{OVO-Bench}
            & \multicolumn{2}{c}{StreamingBench}
            & OVBench \\
            \cmidrule(lr){2-4}
            \cmidrule(lr){5-6}
            \cmidrule(lr){7-7}
            Block & RVP & BT & FAR & RT & PO & Avg. \\
            \midrule
            $B=8$
            & 63.91 & 66.54 & 49.79
            & 68.88 & 52.00 & 55.90 \\
            $B=16$
            & 63.05 & \textbf{69.14} & 52.12
            & 72.28 & 35.20 & 56.95 \\
            $B=32$
            & 62.71 & 67.85 & 53.12
            & 67.64 & \textbf{55.60} & 53.35 \\
            \rowcolor{OursRowBlue}
            $\mathbf{B=4}$
            & \textbf{67.63} & 66.84 & \textbf{53.21}
            & \textbf{73.84} & 31.20 & \textbf{57.16} \\
            \bottomrule
        \end{tabularx}
    \end{minipage}

\end{table*}

\subsection{Ablation Studies}
\label{sec:ablation}

We evaluate the effectiveness of pre-answer gating, blockwise context management, shared context packing, and block size.
Ablation variants of pre-answer gating and block size are evaluated after 4{,}000 training steps for computational efficiency. Additional ablation results are provided in \cref{sec:extended_analysis}.

\begin{figure}[!t]
    \centering
    \includegraphics[width=.97\columnwidth,trim=0bp 6.48bp 0bp 0bp,clip]{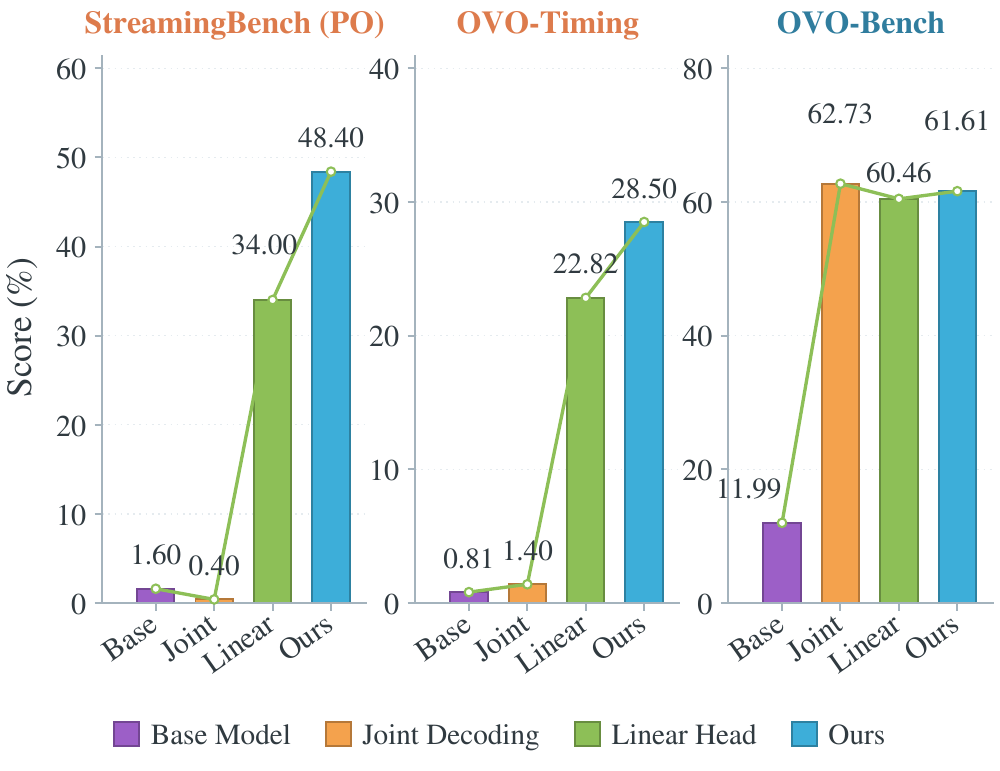}
    \caption{Comparison of response-control designs under a proactive-response protocol with streaming video sampled at 2 FPS. Base Model denotes pretrained Bard-VL-8B for reference.}
    \label{fig:response_control}
\end{figure}

\noindent\textbf{Pre-answer gating.}
To evaluate pre-answer gating (\cref{sec:method_control}), we compare it with two alternative response control designs:
\textit{Joint Decoding}, which places \texttt{<wait>}/\texttt{<answer>} tokens inside the answer block and optimizes them jointly, and \textit{Linear Head}, which predicts response decisions via a separate linear classifier.
As shown in \cref{fig:response_control}, while all trained variants achieve comparable OVO-Bench perception accuracy (60.46--62.73\%), they differ dramatically in proactive response.
Pre-answer gating reaches 48.40\% on StreamingBench-PO and 28.50 on OVO-Timing, outperforming Linear Head (34.00\% and 22.82) and Joint Decoding (0.40\% and 1.40).
This is because the control supervision of Joint Decoding is diluted by answer tokens, whereas Linear Head bypasses the pretrained token prediction head.
Instead, pre-answer gating resolves both issues by executing a gating step through the shared backbone with independent loss normalization.

\noindent\textbf{Blockwise context management.}
To evaluate our context management (\cref{sec:method_architecture}), we compare it with sliding-window baselines that retain only the most recent frames within a 4\,s, 16\,s, or 80\,s window.
Compared with the 4\,s window, our full context management improves overall accuracy from 64.73\% to 68.68\% on OVO-Bench, from 65.67\% to 65.99\% on OVBench, and from 58.40\% to 65.60\% on StreamingBench-PO.
As shown in \cref{tab:temporal_window}, although the 4\,s window achieves higher scores on Real-time Visual Perception (OVO-Bench RVP, 75.00\% \vs 71.10\%) and Spatial Perception (OVBench SP, 66.72\% \vs 60.65\%), it degrades markedly on Forward Active Responding (OVO-Bench FAR, 47.70\% \vs 60.71\%), Future Prediction (OVBench FP, 54.81\% \vs 58.23\%), and Past Memory (OVBench PM, 62.58\% \vs 69.06\%).
By retaining recent frames in full while compressing older segments, blockwise context management effectively balances current-scene perception with long-horizon historical recall.

\noindent\textbf{Shared context packing.}
To evaluate shared context packing (\cref{sec:method_training}), we compare our packed training pipeline against an unpacked baseline where each gating or denoising block processes its historical context in a separate forward pass.
Evaluated on 8 NVIDIA H100 GPUs over 21 batches (672 sessions) that both settings complete without running out of memory, shared context packing reduces the average training time per batch from 136.34\,s to 64.46\,s (a 2.12$\times$ speedup), increases throughput from 0.235 to 0.496 sessions/s, and lowers peak GPU memory from 70.94 to 69.10\,GiB.
These results confirm that packing interactions over a de-duplicated shared KV prefix effectively reduces training redundancy.

\noindent\textbf{Block size.}
We compare StreamDLM-4B variants with block sizes $B\in\{4,8,16,32\}$, initialized from the corresponding Bard-VL-4B checkpoints and trained with the same data and batch settings.
As shown in \cref{tab:block_size}, $B=4$ achieves the best performance on 4 out of 6 metrics, including OVO-Bench RVP and FAR, StreamingBench-RT, and overall OVBench, while $B=16$ and $B=32$ lead on Backward Tracing (OVO-Bench BT) and Proactive Output (StreamingBench-PO), respectively.
Larger blocks allow more tokens to interact bidirectionally within each denoising block, whereas more compact blocks enable more frequent context updates as new video observations arrive.
Since streaming video understanding involves rapidly evolving visual contexts, we adopt $B=4$ as our default to balance intra-block bidirectional interaction with responsive context refreshes.

\begin{table}[!t]
\centering\begin{minipage}{\linewidth}\captionsetup{type=table}
    \centering
    \caption{Inference efficiency comparison on a single NVIDIA A100-SXM4-80GB GPU.}
    \label{tab:inference_efficiency}
    \fontsize{9}{10.5}\selectfont
    \setlength{\tabcolsep}{1.35pt}
    \renewcommand{\arraystretch}{1.11}
    \begin{tabular}{lcrrr}
        \toprule
        Model & Params & TTFT (ms)$\downarrow$ & Memory (GB)$\downarrow$ & TPS$\uparrow$ \\
        \midrule
        VideoChat3 & 4B & 899.3 & 12.1 & 33.0 \\
        Flash-VStream & 7B & 227.0 & 15.2 & 45.6 \\
        ViSpeak & 7B & 204.0 & 18.8 & 29.9 \\
        StreamForest & 7B & 276.4 & 16.9 & 46.8 \\
        Mage-VL & 4B & 334.8 & \underline{11.2} & 32.6 \\
        AURA & 8B & 141.9 & 75.6 & 86.4 \\
        VideoChat-Online & 4B & 147.7 & \textbf{9.8} & 33.1 \\
        VideoLLM-Online & 8B & 126.1 & 19.0 & 25.1 \\
        VST & 7B & 1059.7 & 18.1 & 42.1 \\
        MOSS-VL-Realtime & 11B & 228.7 & 25.1 & 19.0 \\
        Qwen3-VL & 8B & 292.2 & 20.3 & 13.3 \\
        \midrule
        \rowcolor{OursRowBlue}
        \textbf{StreamDLM} & 4B & \textbf{97.5} & 27.1 & \textbf{194.4} \\
        \rowcolor{OursRowBlue}
        \textbf{StreamDLM} & 8B & \underline{100.9} & 43.5 & \underline{139.1} \\
        \bottomrule
    \end{tabular}
\end{minipage}
\end{table}

\subsection{Inference Efficiency}
\label{sec:inference_efficiency}
We benchmark inference efficiency on a single NVIDIA A100-SXM4-80GB GPU using streaming video inputs sampled at 2\,FPS with a per-frame budget of 131{,}072 pixels across four durations (32, 64, 128, and 256\,s).
To control for output length across models, each model generates a standardized 115-word response at the end of each stream.
Tab.~\ref{tab:inference_efficiency} reports the time to first token (TTFT), mean peak GPU memory, and generation throughput in tokens per second (TPS), averaged over three runs per duration and equally across all four durations.

Operating with a block size of $B=4$, StreamDLM-4B and StreamDLM-8B achieve the lowest TTFT (97.5 and 100.9\,ms) and highest throughput (194.4 and 139.1\,TPS) among all evaluated models.

\section{Conclusion}

We introduced StreamDLM, a block-diffusion vision-language model for streaming video understanding.
Pre-answer gating and blockwise context management coordinate response control, intra-block answer denoising, and history updates within a fixed context budget.
Furthermore, shared context packing enables the model to jointly learn response timing and block denoising over causally available history while reducing redundant computation.
Extensive experiments demonstrate state-of-the-art performance in proactive response, real-time interaction, and historical question answering, while achieving lower first-output latency and higher throughput than autoregressive streaming baselines.

\FloatBarrier

{\fontsize{9}{10.48}\selectfont
\setlength{\bibsep}{0pt}
\let\ReportBibitem\bibitem
\renewcommand{\bibitem}[2][]{%
  \ifstrequal{#2}{yao2025timechat}{\balance}{}%
  \ReportBibitem[#1]{#2}}
\bibliographystyle{ieeenat_fullname}
\bibliography{main}

\begin{thebibliography}{83}
\providecommand{\natexlab}[1]{#1}
\providecommand{\url}[1]{\texttt{#1}}
\expandafter\ifx\csname urlstyle\endcsname\relax
  \providecommand{\doi}[1]{doi: #1}\else
  \providecommand{\doi}{doi: \begingroup \urlstyle{rm}\Url}\fi

\bibitem[Anne~Hendricks et~al.(2017)Anne~Hendricks, Wang, Shechtman, Sivic,
  Darrell, and Russell]{anne2017localizing}
Lisa Anne~Hendricks, Oliver Wang, Eli Shechtman, Josef Sivic, Trevor Darrell,
  and Bryan Russell.
\newblock Localizing moments in video with natural language.
\newblock In \emph{Proceedings of the IEEE international conference on computer
  vision}, pages 5803--5812, 2017.

\bibitem[Arriola et~al.(2025)Arriola, Gokaslan, Chiu, Yang, Qi, Han, Sahoo, and
  Kuleshov]{arriola2025block}
Marianne Arriola, Aaron Gokaslan, Justin Chiu, Zhihan Yang, Zhixuan Qi, Jiaqi
  Han, Subham Sahoo, and Volodymyr Kuleshov.
\newblock {Block Diffusion: Interpolating Between Autoregressive and Diffusion
  Language Models}.
\newblock In \emph{Proceedings of the International Conference on Learning
  Representations}, 2025.

\bibitem[Azad et~al.(2026)Azad, Vineet, and Rawat]{azad2026streamready}
Shehreen Azad, Vibhav Vineet, and Yogesh~Singh Rawat.
\newblock {StreamReady: Learning What to Answer and When in Long Streaming
  Videos}.
\newblock In \emph{Proceedings of the IEEE/CVF Conference on Computer Vision
  and Pattern Recognition}. IEEE, 2026.

\bibitem[Bai et~al.(2025{\natexlab{a}})Bai, Cai, Chen, Chen, Chen, Cheng, Deng,
  Ding, Gao, Ge, Ge, Guo, Huang, Huang, Huang, Hui, Jiang, Li, Li, Li, Li, Lin,
  Lin, Liu, Liu, Liu, Liu, Liu, Liu, Lu, Luo, Lv, Men, Meng, Ren, Ren, Song,
  Sun, Tang, Tu, Wan, Wang, Wang, Wang, Wang, Xie, Xu, Xu, Xu, Yang, Yang,
  Yang, Yang, Yu, Zhang, Zhang, Zhang, Zheng, Zhong, Zhou, Zhou, Zhou, Zhu, and
  Zhu]{bai2025qwen3}
Shuai Bai, Yuxuan Cai, Ruizhe Chen, Keqin Chen, Xionghui Chen, Zesen Cheng,
  Lianghao Deng, Wei Ding, Chang Gao, Chunjiang Ge, Wenbin Ge, Zhifang Guo,
  Qidong Huang, Jie Huang, Fei Huang, Binyuan Hui, Shutong Jiang, Zhaohai Li,
  Mingsheng Li, Mei Li, Kaixin Li, Zicheng Lin, Junyang Lin, Xuejing Liu,
  Jiawei Liu, Chenglong Liu, Yang Liu, Dayiheng Liu, Shixuan Liu, Dunjie Lu,
  Ruilin Luo, Chenxu Lv, Rui Men, Lingchen Meng, Xuancheng Ren, Xingzhang Ren,
  Sibo Song, Yuchong Sun, Jun Tang, Jianhong Tu, Jianqiang Wan, Peng Wang,
  Pengfei Wang, Qiuyue Wang, Yuxuan Wang, Tianbao Xie, Yiheng Xu, Haiyang Xu,
  Jin Xu, Zhibo Yang, Mingkun Yang, Jianxin Yang, An Yang, Bowen Yu, Fei Zhang,
  Hang Zhang, Xi Zhang, Bo Zheng, Humen Zhong, Jingren Zhou, Fan Zhou, Jing
  Zhou, Yuanzhi Zhu, and Ke Zhu.
\newblock {Qwen3-VL Technical Report}.
\newblock \emph{arXiv preprint arXiv:2511.21631}, 2025{\natexlab{a}}.

\bibitem[Bai et~al.(2025{\natexlab{b}})Bai, Chen, Liu, Wang, Ge, Song, Dang,
  Wang, Wang, Tang, Zhong, Zhu, Yang, Li, Wan, Wang, Ding, Fu, Xu, Ye, Zhang,
  Xie, Cheng, Zhang, Yang, Xu, and Lin]{bai2025qwen25vl}
Shuai Bai, Keqin Chen, Xuejing Liu, Jialin Wang, Wenbin Ge, Sibo Song, Kai
  Dang, Peng Wang, Shijie Wang, Jun Tang, Humen Zhong, Yuanzhi Zhu, Mingkun
  Yang, Zhaohai Li, Jianqiang Wan, Pengfei Wang, Wei Ding, Zheren Fu, Yiheng
  Xu, Jiabo Ye, Xi Zhang, Tianbao Xie, Zesen Cheng, Hang Zhang, Zhibo Yang,
  Haiyang Xu, and Junyang Lin.
\newblock {Qwen2.5-VL Technical Report}.
\newblock \emph{arXiv preprint arXiv:2502.13923}, 2025{\natexlab{b}}.

\bibitem[Chen et~al.(2026)Chen, Xia, Tu, Shi, Zhang, Yuan, and
  Zhu]{chen2026bard}
Baoyou Chen, Hanchen Xia, Peng Tu, Haojun Shi, Liwei Zhang, Weihao Yuan, and
  Siyu Zhu.
\newblock {BARD: Bridging Autoregressive and Diffusion Vision-Language Models
  via Highly Efficient Progressive Block Merging and Stage-Wise Distillation}.
\newblock In \emph{Proceedings of the ACM International Conference on
  Multimedia}. ACM, 2026.

\bibitem[Chen et~al.(2024)Chen, Lv, Wu, Lin, Song, Gao, Liu, Gao, Mao, and
  Shou]{chen2024videollm}
Joya Chen, Zhaoyang Lv, Shiwei Wu, Kevin~Qinghong Lin, Chenan Song, Difei Gao,
  Jia-Wei Liu, Ziteng Gao, Dongxing Mao, and Mike~Zheng Shou.
\newblock {VideoLLM-online: Online Video Large Language Model for Streaming
  Video}.
\newblock In \emph{Proceedings of the IEEE/CVF Conference on Computer Vision
  and Pattern Recognition}. IEEE, 2024.

\bibitem[Cheng et~al.(2026)Cheng, Jiang, Zhou, Liu, Wang, Zhang, Qi, and
  Zhou]{cheng2026sdarvl}
Shuang Cheng, Yuhua Jiang, Zineng Zhou, Dawei Liu, Tao Wang, Linfeng Zhang,
  Biqing Qi, and Bowen Zhou.
\newblock {SDAR-VL: Stable and Efficient Block-wise Diffusion for
  Vision-Language Understanding}.
\newblock In \emph{Proceedings of the Annual Meeting of the Association for
  Computational Linguistics}, San Diego, California, United States, 2026.
  Association for Computational Linguistics.

\bibitem[Cui et~al.(2026)Cui, Xu, Wang, Yu, Sun, Xu, Wang, He, Ma, Cai,
  et~al.]{cui2026minicpm}
Junbo Cui, Bokai Xu, Chongyi Wang, Tianyu Yu, Weiyue Sun, Yingjing Xu, Tianran
  Wang, Zhihui He, Wenshuo Ma, Tianchi Cai, et~al.
\newblock {MiniCPM-o 4.5: Towards Real-Time Full-Duplex Omni-Modal
  Interaction}.
\newblock \emph{arXiv preprint arXiv:2604.27393}, 2026.

\bibitem[Di et~al.(2025)Di, Yu, Zhang, Li, Zhong, Cheng, Li, He, Shu, and
  Jiang]{di2025rekv}
Shangzhe Di, Zhelun Yu, Guanghao Zhang, Haoyuan Li, Tao Zhong, Hao Cheng, Bolin
  Li, Wanggui He, Fangxun Shu, and Hao Jiang.
\newblock {Streaming Video Question-Answering with In-context Video KV-Cache
  Retrieval}.
\newblock In \emph{Proceedings of the International Conference on Learning
  Representations}, 2025.

\bibitem[Epstein et~al.(2020)Epstein, Chen, and Vondrick]{epstein2020oops}
Dave Epstein, Boyuan Chen, and Carl Vondrick.
\newblock Oops! predicting unintentional action in video.
\newblock In \emph{2020 IEEE/CVF Conference on Computer Vision and Pattern
  Recognition (CVPR)}, pages 916--926. IEEE, 2020.

\bibitem[Fan et~al.(2019)Fan, Lin, Yang, Chu, Deng, Yu, Bai, Xu, Liao, and
  Ling]{fan2019lasot}
Heng Fan, Liting Lin, Fan Yang, Peng Chu, Ge Deng, Sijia Yu, Hexin Bai, Yong
  Xu, Chunyuan Liao, and Haibin Ling.
\newblock {LaSOT: A High-Quality Benchmark for Large-Scale Single Object
  Tracking}.
\newblock In \emph{Proceedings of the IEEE/CVF Conference on Computer Vision
  and Pattern Recognition}. IEEE, 2019.

\bibitem[Fu et~al.(2025)Fu, Yang, Li, Peng, Lin, Wei, Hu, Xie, and
  Zheng]{fu2025vispeak}
Shenghao Fu, Qize Yang, Yuan-Ming Li, Yi-Xing Peng, Kun-Yu Lin, Xihan Wei,
  Jian-Fang Hu, Xiaohua Xie, and Wei-Shi Zheng.
\newblock {ViSpeak: Visual Instruction Feedback in Streaming Videos}.
\newblock In \emph{Proceedings of the IEEE/CVF International Conference on
  Computer Vision}. IEEE, 2025.

\bibitem[Grauman et~al.(2022)Grauman, Westbury, Byrne, Chavis, Furnari,
  Girdhar, Hamburger, Jiang, Liu, Liu, et~al.]{grauman2022ego4d}
Kristen Grauman, Andrew Westbury, Eugene Byrne, Zachary Chavis, Antonino
  Furnari, Rohit Girdhar, Jackson Hamburger, Hao Jiang, Miao Liu, Xingyu Liu,
  et~al.
\newblock {Ego4D}: Around the world in 3,000 hours of egocentric video.
\newblock In \emph{Proceedings of the IEEE/CVF conference on computer vision
  and pattern recognition}, pages 18995--19012, 2022.

\bibitem[Gu et~al.(2018)Gu, Sun, Ross, Vondrick, Pantofaru, Li,
  Vijayanarasimhan, Toderici, Ricco, Sukthankar, et~al.]{gu2018ava}
Chunhui Gu, Chen Sun, David~A Ross, Carl Vondrick, Caroline Pantofaru, Yeqing
  Li, Sudheendra Vijayanarasimhan, George Toderici, Susanna Ricco, Rahul
  Sukthankar, et~al.
\newblock {AVA}: A video dataset of spatio-temporally localized atomic visual
  actions.
\newblock In \emph{2018 IEEE/CVF Conference on Computer Vision and Pattern
  Recognition}, pages 6047--6056. IEEE, 2018.

\bibitem[Guan et~al.(2026)Guan, Yin, Liang, Ju, Luo, Luan, Liu, and
  Bai]{guan2026video}
Yiran Guan, Liang Yin, Dingkang Liang, Jianzhong Ju, Zhenbo Luo, Jian Luan,
  Yuliang Liu, and Xiang Bai.
\newblock {Video Streaming Thinking: VideoLLMs Can Watch and Think
  Simultaneously}.
\newblock In \emph{Proceedings of the European Conference on Computer Vision},
  Cham, 2026. Springer Nature Switzerland.

\bibitem[Han et~al.(2025)Han, Yang, Chang, Yao, and Wang]{han2025shot2story}
Mingfei Han, Linjie Yang, Xiaojun Chang, Lina Yao, and Heng Wang.
\newblock {Shot2Story: A New Benchmark for Comprehensive Understanding of
  Multi-shot Videos}.
\newblock In \emph{Proceedings of the International Conference on Learning
  Representations}, 2025.

\bibitem[He et~al.(2026)He, Chen, Wang, Qin, Shao, Gan, Li, Wu, and
  Lin]{he2026vidlada}
Zhihao He, Tieyuan Chen, Kangyu Wang, Ziran Qin, Yang Shao, Chaofan Gan, Shijie
  Li, Zuxuan Wu, and Weiyao Lin.
\newblock {VidLaDA: Bidirectional Diffusion Large Language Models for Efficient
  Video Understanding}.
\newblock In \emph{Proceedings of the International Conference on Machine
  Learning}. PMLR, 2026.

\bibitem[Hong et~al.(2026)Hong, Yu, Ye, Huang, Zheng, Zhang, Wang, and
  Yao]{hong2026wino}
Feng Hong, Geng Yu, Yushi Ye, Haicheng Huang, Huangjie Zheng, Ya Zhang, Yanfeng
  Wang, and Jiangchao Yao.
\newblock {Wide-In, Narrow-Out: Revokable Decoding for Efficient and Effective
  DLLMs}.
\newblock In \emph{Proceedings of the International Conference on Learning
  Representations}, 2026.

\bibitem[Huang et~al.(2021)Huang, Zhao, and Huang]{huang2019got}
Lianghua Huang, Xin Zhao, and Kaiqi Huang.
\newblock {GOT-10k: A Large High-Diversity Benchmark for Generic Object
  Tracking in the Wild}.
\newblock \emph{IEEE Transactions on Pattern Analysis and Machine
  Intelligence}, 43\penalty0 (5):\penalty0 1562--1577, 2021.

\bibitem[Huang et~al.(2025)Huang, Li, Li, Wang, Zeng, Liang, Wu, Chen, Li, and
  Wang]{huang2025online}
Zhenpeng Huang, Xinhao Li, Jiaqi Li, Jing Wang, Xiangyu Zeng, Cheng Liang, Tao
  Wu, Xi Chen, Liang Li, and Limin Wang.
\newblock {Online Video Understanding: OVBench and VideoChat-Online}.
\newblock In \emph{Proceedings of the IEEE/CVF Conference on Computer Vision
  and Pattern Recognition}. IEEE, 2025.

\bibitem[Kim et~al.(2026)Kim, Lee, Rehg, Kim, and Ro]{kim2026stride}
Junho Kim, Hosu Lee, James~Matthew Rehg, Minsu Kim, and Yong~Man Ro.
\newblock {STRIDE: When to Speak Meets Sequence Denoising for Streaming Video
  Understanding}.
\newblock In \emph{Proceedings of the European Conference on Computer Vision},
  Cham, 2026. Springer Nature Switzerland.

\bibitem[Kim et~al.(2025)Kim, Shim, Choi, and Chang]{kim2025infinipotv}
Minsoo Kim, Kyuhong Shim, Jungwook Choi, and Simyung Chang.
\newblock {InfiniPot-V: Memory-Constrained KV Cache Compression for Streaming
  Video Understanding}.
\newblock In \emph{Advances in Neural Information Processing Systems}. Curran
  Associates, Inc., 2025.

\bibitem[Krishna et~al.(2017)Krishna, Hata, Ren, Fei-Fei, and
  Carlos~Niebles]{krishna2017dense}
Ranjay Krishna, Kenji Hata, Frederic Ren, Li Fei-Fei, and Juan Carlos~Niebles.
\newblock Dense-captioning events in videos.
\newblock In \emph{Proceedings of the IEEE international conference on computer
  vision}, pages 706--715, 2017.

\bibitem[Lei et~al.(2021)Lei, Berg, and Bansal]{lei2021detecting}
Jie Lei, Tamara~L Berg, and Mohit Bansal.
\newblock Detecting moments and highlights in videos via natural language
  queries.
\newblock \emph{Advances in Neural Information Processing Systems},
  34:\penalty0 11846--11858, 2021.

\bibitem[Li et~al.(2025{\natexlab{a}})Li, Zhang, Guo, Zhang, Li, Zhang, Zhang,
  Zhang, Li, Liu, and Li]{li2024llava}
Bo Li, Yuanhan Zhang, Dong Guo, Renrui Zhang, Feng Li, Hao Zhang, Kaichen
  Zhang, Peiyuan Zhang, Yanwei Li, Ziwei Liu, and Chunyuan Li.
\newblock {LLaVA-OneVision: Easy Visual Task Transfer}.
\newblock \emph{Transactions on Machine Learning Research}, 2025{\natexlab{a}}.

\bibitem[Li et~al.(2026{\natexlab{a}})Li, Zhou, Muhtar, Yin, Yan, Shen, Liang,
  Vosoughi, and Liu]{li2026prophet}
Pengxiang Li, Yefan Zhou, Dilxat Muhtar, Lu Yin, Shilin Yan, Li Shen, Yi Liang,
  Soroush Vosoughi, and Shiwei Liu.
\newblock {Diffusion Language Model Knows the Answer Before It Decodes}.
\newblock In \emph{Proceedings of the International Conference on Learning
  Representations}, 2026{\natexlab{a}}.

\bibitem[Li et~al.(2025{\natexlab{b}})Li, Kallidromitis, Bansal, Gokul, Kato,
  Kozuka, Kuen, Lin, Chang, and Grover]{li2025lavida}
Shufan Li, Konstantinos Kallidromitis, Hritik Bansal, Akash Gokul, Yusuke Kato,
  Kazuki Kozuka, Jason Kuen, Zhe Lin, Kai-Wei Chang, and Aditya Grover.
\newblock {LaViDa: A Large Diffusion Language Model for Multimodal
  Understanding}.
\newblock In \emph{Advances in Neural Information Processing Systems}. Curran
  Associates, Inc., 2025{\natexlab{b}}.

\bibitem[Li et~al.(2026{\natexlab{b}})Li, Zhu, Zeng, Dong, Wu, Zhang, Yang, Ma,
  Zhang, Shi, et~al.]{li2026videochat3fullyopenvideo}
Xinhao Li, Yuhan Zhu, Xiangyu Zeng, Yuhao Dong, Haoning Wu, Zhiqiu Zhang,
  Yuandong Yang, Changlian Ma, Qingyu Zhang, Yansong Shi, et~al.
\newblock {VideoChat3: Fully Open Video MLLM for Efficient and Generalist Video
  Understanding}, 2026{\natexlab{b}}.

\bibitem[Lin et~al.(2026)Lin, Fang, Chen, Cheng, Wan, Luo, Wang, Li, Liu, and
  Sun]{lin2026streamingbench}
Junming Lin, Zheng Fang, Chi Chen, Haoxuan Cheng, Zihao Wan, Fuwen Luo, Ziyue
  Wang, Peng Li, Yang Liu, and Maosong Sun.
\newblock {StreamingBench: Assessing the Gap for MLLMs to Achieve Streaming
  Video Understanding}.
\newblock In \emph{Proceedings of the IEEE International Conference on
  Acoustics, Speech, and Signal Processing}. IEEE, 2026.

\bibitem[Liu et~al.(2026)Liu, Shi, Hao, Fu, Bi, Zhang, Hu, Wang, Huang, and
  Liu]{liu2026svcbenchstreamingvideocounting}
Pengyiang Liu, Zhongyue Shi, Hongye Hao, Qi Fu, Xueting Bi, Siwei Zhang,
  Xiaoyang Hu, Zitian Wang, Linjiang Huang, and Si Liu.
\newblock {SVCBench: A Streaming Video Counting Benchmark for Spatial-Temporal
  State Maintenance}.
\newblock In \emph{Proceedings of the European Conference on Computer Vision},
  Cham, 2026. Springer Nature Switzerland.

\bibitem[Lu et~al.(2026)Lu, Bo, Chen, Li, Guo, Guan, Liu, Xu, Sun, Sun, Liu,
  and Li]{lu2026aura}
Xudong Lu, Yang Bo, Jinpeng Chen, Shuhan Li, Xintong Guo, Huankang Guan, Fang
  Liu, Dunyuan Xu, Peiwen Sun, Heyang Sun, Rui Liu, and Hongsheng Li.
\newblock {AURA: Always-On Understanding and Real-Time Assistance via Video
  Streams}.
\newblock \emph{arXiv preprint arXiv:2604.04184}, 2026.

\bibitem[Ma et~al.(2026)Ma, Tang, Guo, Han, Xu, He, Wang, Wang, Chen, Yu, and
  Liu]{ma2026response}
Ke Ma, Jiaqi Tang, Bin Guo, Xueting Han, Ruonan Xu, Qingfeng He, Ziheng Wang,
  Xu Wang, Qifeng Chen, Zhiwen Yu, and Yunhao Liu.
\newblock {Response-G1: Explicit Scene Graph Modeling for Proactive Streaming
  Video Understanding}.
\newblock In \emph{Proceedings of the Annual Meeting of the Association for
  Computational Linguistics}, San Diego, California, United States, 2026.
  Association for Computational Linguistics.

\bibitem[Neupane et~al.(2026)Neupane, Bartaula, Belbase, Adhikari, Ghimire,
  Poudel, Bhattarai, and Paudel]{neupane2026foresight}
Ashok~Prasad Neupane, Dipan Bartaula, Ankit Belbase, Saugat Adhikari, Samip
  Ghimire, Saroj Poudel, Binod Bhattarai, and Danda~Pani Paudel.
\newblock {Foresight: Planning Future Perception in Streaming VLMs without
  Retraining}.
\newblock \emph{arXiv preprint arXiv:2610.03123}, 2026.

\bibitem[Ni et~al.(2026)Ni, Wang, Yue, Yu, Zhao, Hua, Chen, Song, Yu, Zheng,
  and Huang]{ni2026flexibility}
Zanlin Ni, Shenzhi Wang, Yang Yue, Tianyu Yu, Weilin Zhao, Yeguo Hua, Tianyi
  Chen, Jun Song, Cheng Yu, Bo Zheng, and Gao Huang.
\newblock {The Flexibility Trap: Rethinking the Value of Arbitrary Order in
  Diffusion Language Models}.
\newblock In \emph{Proceedings of the International Conference on Machine
  Learning}. PMLR, 2026.

\bibitem[Nie et~al.(2025)Nie, Zhu, You, Zhang, Ou, Hu, Zhou, Lin, Wen, and
  Li]{NEURIPS2025_48b383b2}
Shen Nie, Fengqi Zhu, Zebin You, Xiaolu Zhang, Jingyang Ou, Jun Hu, Jun Zhou,
  Yankai Lin, Ji-Rong Wen, and Chongxuan Li.
\newblock {Large Language Diffusion Models}.
\newblock In \emph{Advances in Neural Information Processing Systems}. Curran
  Associates, Inc., 2025.

\bibitem[Niu et~al.(2025)Niu, Li, Miao, Ge, Zhou, He, Dong, Duan, Ding, Qian,
  Zhang, Zang, Cao, He, and Wang]{niu2025ovo}
Junbo Niu, Yifei Li, Ziyang Miao, Chunjiang Ge, Yuanhang Zhou, Qihao He, Xiaoyi
  Dong, Haodong Duan, Shuangrui Ding, Rui Qian, Pan Zhang, Yuhang Zang, Yuhang
  Cao, Conghui He, and Jiaqi Wang.
\newblock {OVO-Bench: How Far is Your Video-LLMs from Real-World Online Video
  Understanding?}
\newblock In \emph{Proceedings of the IEEE/CVF Conference on Computer Vision
  and Pattern Recognition}. IEEE, 2025.

\bibitem[Oncescu et~al.(2021)Oncescu, Henriques, Liu, Zisserman, and
  Albanie]{oncescu2021queryd}
Andreea-Maria Oncescu, Joao~F Henriques, Yang Liu, Andrew Zisserman, and Samuel
  Albanie.
\newblock {QuerYD}: A video dataset with high-quality text and audio
  narrations.
\newblock In \emph{ICASSP 2021-2021 IEEE International Conference on Acoustics,
  Speech and Signal Processing (ICASSP)}, pages 2265--2269. IEEE, 2021.

\bibitem[{OpenAI}(2026)]{openai2026gpt62astra}
{OpenAI}.
\newblock {GPT-6.2 Astra}.
\newblock OpenAI model documentation, 2026.
\newblock Accessed November 23, 2026.

\bibitem[Patraucean et~al.(2023)Patraucean, Smaira, Gupta, Recasens, Markeeva,
  Banarse, Koppula, Malinowski, Yang, Doersch,
  et~al.]{patraucean2023perception}
Viorica Patraucean, Lucas Smaira, Ankush Gupta, Adria Recasens, Larisa
  Markeeva, Dylan Banarse, Skanda Koppula, Mateusz Malinowski, Yi Yang, Carl
  Doersch, et~al.
\newblock {Perception Test}: A diagnostic benchmark for multimodal video
  models.
\newblock \emph{Advances in Neural Information Processing Systems},
  36:\penalty0 42748--42761, 2023.

\bibitem[Poppi et~al.(2026)Poppi, Uzkent, Garg, Porto, Kessler, Yang, Cornia,
  Baraldi, Cucchiara, and Schiffers]{poppi2026countervid}
Tobia Poppi, Burak Uzkent, Amanmeet Garg, Lucas~Resque Porto, Garin~N Kessler,
  Yezhou Yang, Marcella Cornia, Lorenzo Baraldi, Rita Cucchiara, and Florian
  Schiffers.
\newblock {CounterVid}: counterfactual video generation for mitigating action
  and temporal hallucinations in video-language models.
\newblock In \emph{Proceedings of the 2026 Conference on Empirical Methods in
  Natural Language Processing}, pages 31712--31731, 2026.

\bibitem[Qian et~al.(2025)Qian, Ding, Dong, Zhang, Zang, Cao, Lin, and
  Wang]{qian2025dispiderenablingvideollms}
Rui Qian, Shuangrui Ding, Xiaoyi Dong, Pan Zhang, Yuhang Zang, Yuhang Cao,
  Dahua Lin, and Jiaqi Wang.
\newblock {Dispider: Enabling Video LLMs with Active Real-Time Interaction via
  Disentangled Perception, Decision, and Reaction}.
\newblock In \emph{Proceedings of the IEEE/CVF Conference on Computer Vision
  and Pattern Recognition}. IEEE, 2025.

\bibitem[Regneri et~al.(2013)Regneri, Rohrbach, Wetzel, Thater, Schiele, and
  Pinkal]{regneri2013grounding}
Michaela Regneri, Marcus Rohrbach, Dominikus Wetzel, Stefan Thater, Bernt
  Schiele, and Manfred Pinkal.
\newblock Grounding action descriptions in videos.
\newblock \emph{Transactions of the Association for Computational Linguistics},
  1:\penalty0 25--36, 2013.

\bibitem[Riahi et~al.(2025)Riahi, Radman, Guo, Hedjam, and
  Laaksonen]{riahi2025valor32k}
Ines Riahi, Abduljalil Radman, Zixin Guo, Rachid Hedjam, and Jorma Laaksonen.
\newblock {Valor32k-AVQA v2.0}: Open-ended audio-visual question answering
  dataset and benchmark.
\newblock In \emph{Proceedings of the 33rd ACM International Conference on
  Multimedia}, pages 13097--13103, 2025.

\bibitem[Sigurdsson et~al.(2016)Sigurdsson, Varol, Wang, Farhadi, Laptev, and
  Gupta]{sigurdsson2016hollywood}
Gunnar~A Sigurdsson, G{\"u}l Varol, Xiaolong Wang, Ali Farhadi, Ivan Laptev,
  and Abhinav Gupta.
\newblock Hollywood in homes: Crowdsourcing data collection for activity
  understanding.
\newblock In \emph{European conference on computer vision}, pages 510--526.
  Springer, 2016.

\bibitem[Tang et~al.(2019)Tang, Ding, Rao, Zheng, Zhang, Zhao, Lu, and
  Zhou]{tang2019coin}
Yansong Tang, Dajun Ding, Yongming Rao, Yu Zheng, Danyang Zhang, Lili Zhao,
  Jiwen Lu, and Jie Zhou.
\newblock {COIN}: A large-scale dataset for comprehensive instructional video
  analysis.
\newblock In \emph{2019 IEEE/CVF Conference on Computer Vision and Pattern
  Recognition (CVPR)}, pages 1207--1216. IEEE, 2019.

\bibitem[{Video Understanding Team of JoyAI-VL @ Joy Future Academy,
  JD}(2026)]{yao2026joyaivlinteractionrealtimevisionlanguageinteraction}
{Video Understanding Team of JoyAI-VL @ Joy Future Academy, JD}.
\newblock {JoyAI-VL-Interaction: Real-Time Vision-Language Interaction
  Intelligence}, 2026.

\bibitem[Wang et~al.(2026{\natexlab{a}})Wang, Tan, Zhou, Zhou, Chen, He, Zeng,
  Cheng, Wang, Qian, Wang, Huang, Gao, Huang, Cao, Ran, Liu, Zhu, Wang, Tian,
  Liu, Ye, Wang, Jiang, Feng, Fei, Li, Chen, Gao, Cheng, Li, and
  Qiu]{wang2026mossvl}
Pengyu Wang, Chenkun Tan, Shaojun Zhou, Qirui Zhou, Yanxin Chen, Xingyang He,
  Huazheng Zeng, Jijun Cheng, Chenghao Wang, Xiaomeng Qian, Pengfei Wang, Zhan
  Huang, Shanqing Gao, Wei Huang, Longjun Cao, Wu Ran, Jie Liu, Changtai Zhu,
  Hongkai Wang, Yixian Tian, Chenghao Liu, Zhen Ye, Xinghao Wang, Botian Jiang,
  Guoguo Feng, Zhaoye Fei, Ruixiao Li, Mingshu Chen, Yang Gao, Qinyuan Cheng,
  Shimin Li, and Xipeng Qiu.
\newblock {MOSS-VL Technical Report}, 2026{\natexlab{a}}.

\bibitem[Wang et~al.(2025{\natexlab{a}})Wang, Gao, Gu, Pu, Cui, Wei, Liu, Jing,
  Ye, Shao, et~al.]{internvl2025internvl35}
Weiyun Wang, Zhangwei Gao, Lixin Gu, Hengjun Pu, Long Cui, Xingguang Wei,
  Zhaoyang Liu, Linglin Jing, Shenglong Ye, Jie Shao, et~al.
\newblock {InternVL3.5: Advancing Open-Source Multimodal Models in Versatility,
  Reasoning, and Efficiency}.
\newblock \emph{arXiv preprint arXiv:2508.18265}, 2025{\natexlab{a}}.

\bibitem[Wang et~al.(2026{\natexlab{b}})Wang, Fang, Jing, Shen, Shen, Wang,
  Ouyang, Chen, and Shen]{wang2026timefeature}
Wen Wang, Bozhen Fang, Chenchen Jing, Yongliang Shen, Yangyi Shen, Qiuyu Wang,
  Hao Ouyang, Hao Chen, and Chunhua Shen.
\newblock {Time Is a Feature: Exploiting Temporal Dynamics in Diffusion
  Language Models}.
\newblock In \emph{Proceedings of the International Conference on Learning
  Representations}, 2026{\natexlab{b}}.

\bibitem[Wang et~al.(2025{\natexlab{b}})Wang, Meng, Wang, Liang, Wei, Zhang,
  and Zhao]{wang2025videollm}
Yueqian Wang, Xiaojun Meng, Yuxuan Wang, Jianxin Liang, Jiansheng Wei, Huishuai
  Zhang, and Dongyan Zhao.
\newblock {VideoLLM} knows when to speak: Enhancing time-sensitive video
  comprehension with video-text duet interaction format.
\newblock In \emph{EMNLP (Findings)}, pages 6338--6359, 2025{\natexlab{b}}.

\bibitem[Wang et~al.(2025{\natexlab{c}})Wang, Meng, Wang, Zhang, and
  Zhao]{wang2025proactivevideoqa}
Yueqian Wang, Xiaojun Meng, Yifan Wang, Huishuai Zhang, and Dongyan Zhao.
\newblock {ProactiveVideoQA: A Comprehensive Benchmark Evaluating Proactive
  Interactions in Video Large Language Models}.
\newblock \emph{arXiv preprint arXiv:2507.09313}, 2025{\natexlab{c}}.

\bibitem[Wei et~al.(2026)Wei, Zhang, Liu, Zeng, Wang, Qi, Liu, and
  Zhang]{wei2026slowfast}
Qingyan Wei, Yaojie Zhang, Zhiyuan Liu, Puyu Zeng, Yuxuan Wang, Biqing Qi,
  Dongrui Liu, and Linfeng Zhang.
\newblock {Accelerating Diffusion Large Language Models with SlowFast Sampling:
  The Three Golden Principles}.
\newblock In \emph{Proceedings of the International Conference on Learning
  Representations}, 2026.

\bibitem[Wu et~al.(2021)Wu, Yu, Chen, Tenenbaum, and Gan]{wu2024star}
Bo Wu, Shoubin Yu, Zhenfang Chen, Joshua~B. Tenenbaum, and Chuang Gan.
\newblock {STAR: A Benchmark for Situated Reasoning in Real-World Videos}.
\newblock In \emph{Proceedings of the Neural Information Processing Systems
  Track on Datasets and Benchmarks}, 2021.

\bibitem[Wu et~al.(2026{\natexlab{a}})Wu, Lan, Fu, Gao, Wang, Yu, Álvarez,
  Molchanov, Luo, Han, Zhu, and Xie]{wu2026fastdvlm}
Chengyue Wu, Shiyi Lan, Yonggan Fu, Sensen Gao, Jin Wang, Jincheng Yu,
  José~Manuel Álvarez, Pavlo Molchanov, Ping Luo, Song Han, Ligeng Zhu, and
  Enze Xie.
\newblock {Fast-dVLM: Efficient Block-Diffusion VLM via Direct Conversion from
  Autoregressive VLM}.
\newblock \emph{arXiv preprint arXiv:2604.06832}, 2026{\natexlab{a}}.

\bibitem[Wu et~al.(2026{\natexlab{b}})Wu, Zhang, Xue, Diao, Fu, Liu, Molchanov,
  Luo, Han, and Xie]{wu2026fastdllmv2}
Chengyue Wu, Hao Zhang, Shuchen Xue, Shizhe Diao, Yonggan Fu, Zhijian Liu,
  Pavlo Molchanov, Ping Luo, Song Han, and Enze Xie.
\newblock {Fast-dLLM v2: Efficient Block-Diffusion LLM}.
\newblock In \emph{Proceedings of the International Conference on Learning
  Representations}, 2026{\natexlab{b}}.

\bibitem[Wu et~al.(2026{\natexlab{c}})Wu, Zhang, Xue, Liu, Diao, Zhu, Luo, Han,
  and Xie]{wu2026fast}
Chengyue Wu, Hao Zhang, Shuchen Xue, Zhijian Liu, Shizhe Diao, Ligeng Zhu, Ping
  Luo, Song Han, and Enze Xie.
\newblock {Fast-dLLM: Training-free Acceleration of Diffusion LLM by Enabling
  KV Cache and Parallel Decoding}.
\newblock In \emph{Proceedings of the International Conference on Learning
  Representations}, 2026{\natexlab{c}}.

\bibitem[Xiao et~al.(2021)Xiao, Shang, Yao, and Chua]{xiao2021next}
Junbin Xiao, Xindi Shang, Angela Yao, and Tat-Seng Chua.
\newblock {NExT-QA}: Next phase of question-answering to explaining temporal
  actions.
\newblock In \emph{2021 IEEE/CVF Conference on Computer Vision and Pattern
  Recognition (CVPR)}, pages 9772--9781. IEEE, 2021.

\bibitem[Xie et~al.(2024)Xie, Zhang, Zhou, Li, Zhang, Hessel, Yang, and
  Liu]{xie2024funqa}
Binzhu Xie, Sicheng Zhang, Zitang Zhou, Bo Li, Yuanhan Zhang, Jack Hessel,
  Jingkang Yang, and Ziwei Liu.
\newblock {FunQA}: Towards surprising video comprehension.
\newblock In \emph{European Conference on Computer Vision}, pages 39--57.
  Springer, 2024.

\bibitem[Xu et~al.(2026)Xu, Xiao, Chen, He, Peng, Lu, and
  Han]{xu2026streamingvlm}
Ruyi Xu, Guangxuan Xiao, Yukang Chen, Liuning He, Kelly Peng, Yao Lu, and Song
  Han.
\newblock {StreamingVLM: Real-Time Understanding for Infinite Video Streams}.
\newblock In \emph{Proceedings of the International Conference on Learning
  Representations}, 2026.

\bibitem[Xun et~al.(2025)Xun, Tao, Li, Shi, Lin, Zhu, Yan, Li, Zhang, Wang,
  Liu, Zhang, Ma, and Hu]{xun2026rtvbenchbenchmarkingmllmcontinuous}
Shuhang Xun, Sicheng Tao, Jungang Li, Yibo Shi, Zhixin Lin, Zhanhui Zhu, Yibo
  Yan, Hanqian Li, Linghao Zhang, Shikang Wang, Yixin Liu, Hanbo Zhang, Ying
  Ma, and Xuming Hu.
\newblock {RTV-Bench: Benchmarking MLLM Continuous Perception, Understanding
  and Reasoning through Real-Time Video}.
\newblock In \emph{Advances in Neural Information Processing Systems}. Curran
  Associates, Inc., 2025.

\bibitem[Yan et~al.(2025)Yan, Xu, Di, Liu, Shi, Chen, Li, Huang, and
  Xie]{yan2025streamformer}
Yibin Yan, Jilan Xu, Shangzhe Di, Yikun Liu, Yudi Shi, Qirui Chen, Zeqian Li,
  Yifei Huang, and Weidi Xie.
\newblock {Learning Streaming Video Representation via Multitask Training}.
\newblock In \emph{Proceedings of the IEEE/CVF International Conference on
  Computer Vision}. IEEE, 2025.

\bibitem[Yang et~al.(2025{\natexlab{a}})Yang, Yang, Huang, Brown, Yang, Yu,
  Tong, Zheng, Xu, Wang, Lu, Fergus, LeCun, Fei-Fei, and Xie]{yang2025cambrian}
Shusheng Yang, Jihan Yang, Pinzhi Huang, Ellis Brown, Zihao Yang, Yue Yu,
  Shengbang Tong, Zihan Zheng, Yifan Xu, Muhan Wang, Danhao Lu, Rob Fergus,
  Yann LeCun, Li Fei-Fei, and Saining Xie.
\newblock {Cambrian-S}: Towards spatial supersensing in video.
\newblock \emph{arXiv preprint arXiv:2511.04670}, 2025{\natexlab{a}}.

\bibitem[Yang et~al.(2026)Yang, Zhang, Jia, Guo, Shen, Zhang, Zhang, Wang, Li,
  Zhang, et~al.]{yang2026magevlefficientcodecnativestreaming}
Senqiao Yang, Kaichen Zhang, Zhaoyang Jia, Jinghao Guo, Yifei Shen, Xinjie
  Zhang, Xiaoyi Zhang, Haoqing Wang, Xiao Li, Peng Zhang, et~al.
\newblock {Mage-VL: An Efficient Codec-Native Streaming Multimodal Foundation
  Model}.
\newblock arXiv preprint arXiv:2607.24904, 2026.

\bibitem[Yang et~al.(2025{\natexlab{b}})Yang, Zhao, Shukla, Singh, Mishra,
  Zhang, and Ren]{yang2025streammem}
Yanlai Yang, Zhuokai Zhao, Satya~Narayan Shukla, Aashu Singh, Shlok~Kumar
  Mishra, Lizhu Zhang, and Mengye Ren.
\newblock {StreamMem: Query-Agnostic KV Cache Memory for Streaming Video
  Understanding}.
\newblock arXiv preprint arXiv:2508.15717, 2025{\natexlab{b}}.

\bibitem[Yao et~al.(2025)Yao, Li, Wei, Li, Ren, Liu, Ouyang, Wang, Li, Li,
  Kong, Liu, Zhang, and Sun]{yao2025timechat}
Linli Yao, Yicheng Li, Yuancheng Wei, Lei Li, Shuhuai Ren, Yuanxin Liu, Kun
  Ouyang, Lean Wang, Shicheng Li, Sida Li, Lingpeng Kong, Qi Liu, Yuanxing
  Zhang, and Xu Sun.
\newblock {TimeChat-Online: 80\% Visual Tokens are Naturally Redundant in
  Streaming Videos}.
\newblock In \emph{Proceedings of the ACM International Conference on
  Multimedia}. ACM, 2025.

\bibitem[Ye et~al.(2025{\natexlab{a}})Ye, Gong, Gao, Fan, Wu, Bi, Bai, Shang,
  and Kong]{ye2025dreamvl}
Jiacheng Ye, Shansan Gong, Jiahui Gao, Junming Fan, Shuang Wu, Wei Bi, Haoli
  Bai, Lifeng Shang, and Lingpeng Kong.
\newblock {Dream-VL \& Dream-VLA: Open Vision-Language and
  Vision-Language-Action Models with Diffusion Language Model Backbone},
  2025{\natexlab{a}}.

\bibitem[Ye et~al.(2025{\natexlab{b}})Ye, Xie, Zheng, Gao, Wu, Jiang, Li, and
  Kong]{ye2508dream}
Jiacheng Ye, Zhihui Xie, Lin Zheng, Jiahui Gao, Zirui Wu, Xin Jiang, Zhenguo
  Li, and Lingpeng Kong.
\newblock {Dream 7B: Diffusion Large Language Models}.
\newblock arXiv preprint arXiv:2508.15487, 2025{\natexlab{b}}.

\bibitem[Yoon et~al.(2025)Yoon, Yoon, Hasegawa-Johnson, and Yoo]{yoon2025can}
Eunseop Yoon, Hee~Suk Yoon, Mark Hasegawa-Johnson, and Chang Yoo.
\newblock Can video {LLMs} refuse to answer? alignment for answerability in
  video large language models.
\newblock In \emph{International Conference on Learning Representations}, pages
  86500--86525, 2025.

\bibitem[You et~al.(2026)You, Nie, Zhang, Zhou, Lu, Wen, and Li]{you2026llada}
Zebin You, Shen Nie, Xiaolu Zhang, Jun Zhou, Zhiwu Lu, Ji-Rong Wen, and
  Chongxuan Li.
\newblock {LLaDA-V: Large Language Diffusion Models with Visual Instruction
  Tuning}.
\newblock In \emph{Proceedings of the IEEE/CVF Conference on Computer Vision
  and Pattern Recognition}. IEEE, 2026.

\bibitem[Yu et~al.(2026)Yu, Ma, and Wang]{yu2025dimple}
Runpeng Yu, Xinyin Ma, and Xinchao Wang.
\newblock {Dimple: Discrete Diffusion Parallel Generation for Large Multimodal
  Modal}.
\newblock In \emph{Proceedings of the Conference on Empirical Methods in
  Natural Language Processing}, Budapest, Hungary, 2026. Association for
  Computational Linguistics.

\bibitem[Yu et~al.(2019)Yu, Xu, Yu, Yu, Zhao, Zhuang, and
  Tao]{yu2019activitynet}
Zhou Yu, Dejing Xu, Jun Yu, Ting Yu, Zhou Zhao, Yueting Zhuang, and Dacheng
  Tao.
\newblock {ActivityNet-QA}: A dataset for understanding complex web videos via
  question answering.
\newblock In \emph{Proceedings of the AAAI conference on artificial
  intelligence}, pages 9127--9134, 2019.

\bibitem[Zadeh et~al.(2019)Zadeh, Chan, Liang, Tong, and
  Morency]{zadeh2019social}
Amir Zadeh, Michael Chan, Paul~Pu Liang, Edmund Tong, and Louis-Philippe
  Morency.
\newblock {Social-IQ}: A question answering benchmark for artificial social
  intelligence.
\newblock In \emph{2019 IEEE/CVF Conference on Computer Vision and Pattern
  Recognition (CVPR)}, pages 8799--8809. IEEE, 2019.

\bibitem[Zeng et~al.(2026{\natexlab{a}})Zeng, Yao, Liao, Tao, Liu, and
  Wang]{zeng2025diffusionvl}
Lunbin Zeng, Jingfeng Yao, Bencheng Liao, Hongyuan Tao, Wenyu Liu, and Xinggang
  Wang.
\newblock {DiffusionVL: Translating Any Autoregressive Models into Diffusion
  Vision Language Models}.
\newblock In \emph{Proceedings of the European Conference on Computer Vision},
  Cham, 2026{\natexlab{a}}. Springer Nature Switzerland.

\bibitem[Zeng et~al.(2025)Zeng, Qiu, Zhang, Li, Wang, Li, Yan, Tian, Tian,
  Zhao, Wang, and Wang]{zeng2026streamforest}
Xiangyu Zeng, Kefan Qiu, Qingyu Zhang, Xinhao Li, Jing Wang, Jiaxin Li, Ziang
  Yan, Kun Tian, Meng Tian, Xinhai Zhao, Yi Wang, and Limin Wang.
\newblock {StreamForest: Efficient Online Video Understanding with Persistent
  Event Memory}.
\newblock In \emph{Advances in Neural Information Processing Systems}. Curran
  Associates, Inc., 2025.

\bibitem[Zeng et~al.(2026{\natexlab{b}})Zeng, Yang, Zhang, Zhu, Li, Si, Yao,
  Ma, Chen, Chen, Shi, Zhou, Li, Zhang, Qin, Yang, Yu, Ouyang, Shao, Wei, Zhou,
  Gao, Wang, and Wang]{zeng2026onestreamer}
Xiangyu Zeng, Yuandong Yang, Zhiqiu Zhang, Yuhan Zhu, Xinhao Li, Qingyi Si,
  Dingyu Yao, Changlian Ma, Haoran Chen, Xinyu Chen, Yansong Shi, Junhao Zhou,
  Yifei Li, Jun Zhang, Chuanyu Qin, Chenxu Yang, Xinlei Yu, Kun Ouyang, Yuchen
  Shao, Qianshan Wei, Changhai Zhou, Jun Gao, Jiaqi Wang, and Limin Wang.
\newblock {OneStreamer: Unifying Perception, Memory, and Proactive Response in
  Streaming Video Interaction}.
\newblock \emph{arXiv preprint arXiv:2610.01762}, 2026{\natexlab{b}}.

\bibitem[Zeng et~al.(2026{\natexlab{c}})Zeng, Zhang, Zhu, Li, Wang, Ma, Zhang,
  Huang, Ouyang, Jiang, et~al.]{zeng2026video}
Xiangyu Zeng, Zhiqiu Zhang, Yuhan Zhu, Xinhao Li, Zikang Wang, Changlian Ma,
  Qingyu Zhang, Zizheng Huang, Kun Ouyang, Tianxiang Jiang, et~al.
\newblock {Video-o3}: Native interleaved clue seeking for long video multi-hop
  reasoning.
\newblock \emph{arXiv preprint arXiv:2601.23224}, 2026{\natexlab{c}}.

\bibitem[Zhang et~al.(2025{\natexlab{a}})Zhang, Wang, Tang, Liu, Feng, and
  Jin]{zhang2024flash}
Haoji Zhang, Yiqin Wang, Yansong Tang, Yong Liu, Jiashi Feng, and Xiaojie Jin.
\newblock {Flash-VStream: Efficient Real-Time Understanding for Long Video
  Streams}.
\newblock In \emph{Proceedings of the IEEE/CVF International Conference on
  Computer Vision}. IEEE, 2025{\natexlab{a}}.

\bibitem[Zhang et~al.(2025{\natexlab{b}})Zhang, Lu, Hu, Wang, Yuan, Liu, and
  Wu]{zhang2025period}
Yuting Zhang, Hao Lu, Qingyong Hu, Yin Wang, Kaishen Yuan, Xin Liu, and Kaishun
  Wu.
\newblock {Period-LLM}: Extending the periodic capability of multimodal large
  language model.
\newblock In \emph{2025 IEEE/CVF Conference on Computer Vision and Pattern
  Recognition (CVPR)}, pages 29237--29247. IEEE, 2025{\natexlab{b}}.

\bibitem[Zhang et~al.(2025{\natexlab{c}})Zhang, Wu, Li, Li, Ma, Liu, and
  Li]{zhang2024llava}
Yuanhan Zhang, Jinming Wu, Wei Li, Bo Li, Zejun Ma, Ziwei Liu, and Chunyuan Li.
\newblock {LLaVA-Video: Video Instruction Tuning With Synthetic Data}.
\newblock \emph{Transactions on Machine Learning Research}, 2025{\natexlab{c}}.

\bibitem[Zhao et~al.(2026)Zhao, Wang, Li, Qian, Lin, and
  Liu]{zhao2026cogstream}
Zicheng Zhao, Kangyu Wang, Shijie Li, Rui Qian, Weiyao Lin, and Huabin Liu.
\newblock {CogStream}: Context-guided streaming video question answering.
\newblock In \emph{Proceedings of the AAAI Conference on Artificial
  Intelligence}, pages 13332--13341, 2026.

\bibitem[Zheng et~al.(2026)Zheng, Ding, Yang, Jiang, Wu, Zhang, Wang, Cao, and
  Liu]{zheng2026emgardeproposematchframeworkproactive}
Yikai Zheng, Xin Ding, Yifan Yang, Shiqi Jiang, Hao Wu, Qianxi Zhang, Weijun
  Wang, Ting Cao, and Yunxin Liu.
\newblock {Em-Garde: A Propose-Match Framework for Proactive Streaming Video
  Understanding}.
\newblock In \emph{Proceedings of the European Conference on Computer Vision},
  Cham, 2026. Springer Nature Switzerland.

\bibitem[Zhou et~al.(2018)Zhou, Xu, and Corso]{zhou2018towards}
Luowei Zhou, Chenliang Xu, and Jason Corso.
\newblock Towards automatic learning of procedures from web instructional
  videos.
\newblock In \emph{Proceedings of the AAAI conference on artificial
  intelligence}, 2018.

\end{thebibliography}
}

\clearpage
\nobalance
\flushbottom
\appendix
\setcounter{table}{0}
\setcounter{figure}{0}
\setcounter{equation}{0}
\renewcommand{\thetable}{S\arabic{table}}
\renewcommand{\thefigure}{S\arabic{figure}}
\renewcommand{\theequation}{S\arabic{equation}}
\renewcommand{\theHtable}{supp.\arabic{table}}
\renewcommand{\theHfigure}{supp.\arabic{figure}}
\renewcommand{\theHequation}{supp.\arabic{equation}}
\twocolumn[{%
\phantomsection\pdfbookmark[0]{Supplementary Material}{supplement}%
\begin{center}
  {\large\bfseries StreamDLM: Streaming Video Understanding\\
  with Diffusion Large Language Model\par}
  \vspace{4pt}
  {\Large\bfseries Supplementary Material\par}
\end{center}
\vspace{6pt}
}]
\section{Method and Implementation Details}
\label{sec:supp_method}

% =========================================================================
\subsection{StreamDLM Architecture (\S3.1)}
\label{sec:supp_arch}

\noindent \textbf{Decoding a four-token block.}
Each answer block starts with four masked positions. At each denoising step, the model uses the same context snapshot $\mathcal C_b$ to predict a token at every position that is still masked. It accepts all predictions whose probability exceeds $0.5$. If none exceed the threshold, it accepts the single most confident prediction. Equivalently, if $U_r$ is the set of masked positions at step $r$, $c_i$ is the probability of the most likely token at position $i$, and $i^\star$ is a position with the highest such probability in $U_r$, the accepted positions are
\begin{equation}
T_r = \{i \in U_r : c_i > 0.5\} \cup \{i^\star\}.
\label{eq:supp_unmask}
\end{equation}
Accepted tokens stay fixed. Therefore, each step fills at least one position, and decoding stops as soon as all four are filled, after at most four steps.

\noindent \textbf{Sampling and encoding incoming video.}
In the default setting, we sample one frame every 0.5\,s. Once per second, we feed the two sampled frames to the visual encoder together and append the resulting segment to the observation history $\mathcal H_o$. For each sampling time $t$, we select the source frame with the largest timestamp no greater than $t$. For example, if consecutive frames are timestamped 0.46 and 0.51\,s, the sample at 0.50\,s uses the frame at 0.46\,s: the frame at 0.51\,s has not yet arrived. If only one frame is sampled, we duplicate it to form the visual encoder's required pair.

To add the new segment to the observation history $\mathcal H_o$, we divide its input tokens into blocks of four. A token can attend to all cached observation tokens, every earlier block in the segment, and every token in its own block. It cannot attend to later blocks. Let $H$ denote the cached token positions and $B_1,B_2,\ldots$ the sets of positions in the new input blocks. For a token $i$ in block $B_b$, the attention mask is
\begin{equation}
M^{\mathrm{src}}_{i,j}
= \mathbf{1}\!\left[j\in H\cup B_1\cup\cdots\cup B_b\right].
\label{eq:supp_source_mask}
\end{equation}

\noindent \textbf{Assigning positions and reading the cache.}
Incoming video and generated text share one position counter, $p_{\mathrm{next}}$. Before decoding a four-token answer block, the model reserves the next four positions for it. Frames that arrive while the block is being decoded receive later positions. For 3D M-RoPE, the three coordinates advance together for text tokens. Within a video frame, visual tokens share the temporal coordinate at the frame's start; their other two coordinates specify the row and column in the merged spatial grid.

At the start of block $b$, the decoder reads the latest observation history and the retained answer tokens committed before that block. Its snapshot $\mathcal C_b$ references the stored KV states for both histories without copying them. Frames arriving during denoising reach the next block; the current view stays fixed.

% =========================================================================
\subsection{Pre-Answer Gating (\S3.2)}
\label{sec:supp_gating}

\noindent \textbf{Deciding whether to speak.}
Before generating any answer tokens, the model reads the latest observations and evaluates the control block $[\texttt{<wait>},\texttt{<answer>},\texttt{[MASK]}]$. The first two tokens represent the choices: keep observing or start an answer. At the third position, the output head maps the hidden state of \texttt{[MASK]} to one score per action. The gate selects the action with the higher score. The block uses three temporary positions starting at $p_{\mathrm{next}}$ without advancing the position counter.

On \texttt{WAIT}, the model continues observing. On \texttt{ANSWER}, the first answer block reuses the gating snapshot (\(\mathcal{C}_1=\mathcal{C}\)). Each subsequent block takes a fresh snapshot of the retained observation and answer histories. The first \texttt{[EOS]} token in the finalized output ends the utterance, not the question: the gate can answer the same question again using newly arrived frames. During answer generation, the logits of both control tokens are set to $-\infty$.

% =========================================================================
\subsection{Blockwise Context Management (\S3.3)}
\label{sec:supp_context}

\noindent \textbf{Committing an answer block.}
After decoding block $\mathbf y_b$, we keep its tokens through the first \texttt{[EOS]} token, or the entire block if it contains no \texttt{[EOS]} token. A clean forward pass on $\mathcal C_b$ commits their KV states to the answer history $\mathcal H_a$. If generation continues, the next block reads the updated answer history and latest observations.

\noindent \textbf{Retaining older visual observations.}
When a segment leaves the 80-second window, its question stays in the observation history as text. We retain all its visual tokens if they fit in long-term memory. Otherwise, we sample visual groups uniformly over time and keep each group's timestamp and some spatial tokens.

\noindent \textbf{Merging visual segments and evicting text.}
When the stored segments together exceed the long-term budget, we compare neighboring segments to decide which pair to merge. We represent each segment $c$ by averaging its raw last-layer value vectors $\mathbf w_i$ over its visual positions $V(c)$ and normalizing the result:
\begin{equation}
\boldsymbol\mu_c=\frac{1}{|V(c)|}\sum_{i\in V(c)}\mathbf w_i,
\qquad
\bar{\mathbf u}_c=\frac{\boldsymbol\mu_c}{\|\boldsymbol\mu_c\|_2}.
\label{eq:supp_chunk_sim}
\end{equation}
We merge the pair with the largest $\bar{\mathbf u}_c^\top\bar{\mathbf u}_{c+1}$ and compress the result if needed. If a single segment exceeds the budget, we compress it directly.
Each cached answer points to its question. We protect the initial sink tokens, the active question, and all questions with cached answers. Remaining space goes to the newest unpaired text. If the protected text exceeds its budget or the answer cache exceeds its own, we evict the oldest completed answer. Its question can be removed only if it is inactive and no retained answer still refers to it.

% =========================================================================
\subsection{Shared Context Packing (\S3.4)}
\label{sec:supp_packing}
\label{sec:supp_training_details}

We replay each training session in chronological order. An \emph{event} is one replay entry: it introduces a new observation or reuses the current one, together with any question and supervision attached to that entry. A second question at that timestamp can form a new event without a new frame. Observations and questions update $\mathcal H_o$; reference answers update $\mathcal H_a$. At an annotated decision, we save the current context with its \texttt{WAIT} or \texttt{ANSWER} label. The decision updates neither history.

\noindent\textbf{Source packing.} When the visible histories of consecutive observations are known, one encoder pass replaces a separate pass per observation. Using token positions and cached values, we track which observations leave the window and which text is evicted. Each segment reads only what was visible at its arrival (Eq.~\eqref{eq:supp_source_mask}). We close the group before the next entry exceeds a token, mask, or cache budget or requires a feature not yet computed. We commit KV states in arrival order.

\noindent\textbf{Clean KV packing.} Training provides the reference answer. When consecutive blocks fit without eviction, we compute their clean KV states in one parallel pass. The block-causal mask lets each block read earlier blocks and itself. We commit the states in order (main paper, Eq.~(6)); denoising uses each block's snapshot \emph{before} it is cached.

\noindent\textbf{Query packing.} A training query predicts a gate action or the masked tokens of an answer block. We batch these predictions into one pass and store their shared KV history only once in $\mathcal P$, avoiding a separate prefix copy per query. Each query reads its own block and only its saved portion of $\mathcal P$ (main paper, Eq.~(7)). Two gate decisions can share past frames; only the later one sees frames arriving between them. We mask answer blocks independently at rates $\rho_b\in(0,1]$, excluding positions after the \texttt{[EOS]} token, and evaluate queued queries before compaction replaces a shared entry.

\noindent\textbf{Segment-wise truncated backpropagation.}
We group consecutive events into \emph{backpropagation segments} and accumulate the loss within each segment before running backward. Each supervised segment contains at most 16 events. Events need not cost the same: a long answer or a long prefix requires more attention work. We estimate the cost of event $e$ as the number of token positions it processes, $Q(e)$, times the estimated number of keys per position:
\begin{equation}
W(e)=Q(e)\min\!\left\{L_{\max},\,L_{\mathrm{pref}}(e)+Q(e)\right\}.
\label{eq:supp_workload}
\end{equation}
Here $L_{\mathrm{pref}}(e)$ estimates the cached prefix length and $L_{\max}$ denotes the context length limit. The count $Q(e)$ includes new inputs, control queries, clean answer writes, and supervised denoising reads. We start a new segment if adding the next event would take the accumulated cost above $1.2\times10^7$; an event that exceeds this budget on its own runs alone. Where the budgets permit, we keep annotated evidence and its question in the same segment so gradients reach the evidence. Unsupervised segments run without gradient tracking. At a segment boundary, we finish pending queries, backpropagate the loss, and detach the retained KV states. This releases the completed segment's computation graph while preserving its KV values and positions for the next segment.

% =========================================================================
\subsection{Training Objective (\S3.5)}
\label{sec:supp_objective}

We accumulate the loss one segment at a time, using normalization counts from the full session (main paper, Eqs.~(8)--(9)). Let $\ell_{\mathrm{ct},m}$ and $\ell_{\mathrm{dn},m}$ be the contributions from segment $m$. For a session with $M$ segments, the training loss is
\begin{equation}
\mathcal L = \frac{1}{n_{\mathrm{obj}}}
\sum_{m=1}^{M}\bigl(\ell_{\mathrm{ct},m}+\ell_{\mathrm{dn},m}\bigr).
\label{eq:supp_session_sum}
\end{equation}
When both objectives have supervision, $n_{\mathrm{obj}}=2$ and each receives half the weight. With one supervised objective, $n_{\mathrm{obj}}=1$; the other contributes zero.

Each segment sums only its own decisions and masked tokens. For control, we use the session-wide counts of \texttt{WAIT} and \texttt{ANSWER}, giving equal weight to each action class present. For denoising, we retain the inverse masking-rate weight $1/\rho_b$, normalize by each answer's valid length including \texttt{[EOS]}, and average over all supervised answers in the session. Thus splitting a session changes neither the weight of a decision nor that of a token. Detaching KV states at segment boundaries truncates gradients; model parameters are updated only after gradient accumulation.

% =========================================================================
\section{Training Data and Configuration}
\label{sec:supp_training}
\label{sec:supp_training_setup}

This section supplements \S4.1 of the main paper with the training data composition and optimization settings for each model size.

% -------------------------------------------------------------------------
\subsection{Training Data Composition}
\label{sec:supp_data}
\label{sec:supp_dataset_statistics}

StreamDLM-4B and StreamDLM-8B use the same training dataset. We also performed train--test deduplication to remove overlapping samples. Table~\ref{tab:supp_data_composition} breaks down the 598,400 sessions reported in \S4.1 of the main paper. We report the cumulative session counts through training step 18,700. Of these, 219,379 (36.66\%) contain at least one \texttt{WAIT} or \texttt{ANSWER} label; the remaining 379,021 (63.34\%) provide answer-content supervision only. Four citations identify the annotation sources used for particular rows: ActivityNet uses the ActivityNet Captions event annotations; CIRCLE-40K draws its annotations from the VSI-590K corpus released with Cambrian-S; YouTube-Collected uses the self-collected portion of Seeker-173K (Video-o3); and Countix uses the Countix-QA release of Period-LLM.

\begin{table}[!t]
\centering
\caption{Training sessions by source. All percentages use the same denominator: 598,400 sessions.}
\label{tab:supp_data_composition}
\label{tab:training_data_composition}
\suppSetupTableStyle
\setlength{\tabcolsep}{2.5pt}
\begin{tabularx}{\linewidth}{@{}l>{\raggedleft\arraybackslash}Xrr@{}}
\toprule
\textbf{Dataset / Source} & \textbf{Control (\%)} & \textbf{Content (\%)} & \textbf{Total (\%)} \\
\midrule
LLaVA-Video~\cite{zhang2024llava} & 15.50 & 1.67 & 17.17 \\
LaSOT~\cite{fan2019lasot} & 0.00 & 11.00 & 11.00 \\
GOT-10k~\cite{huang2019got} & 0.00 & 7.65 & 7.65 \\
Shot2Story~\cite{han2025shot2story} & 0.00 & 7.21 & 7.21 \\
TimeChat-Online~\cite{yao2025timechat} & 0.00 & 6.23 & 6.23 \\
STAR~\cite{wu2024star} & 0.00 & 5.31 & 5.31 \\
MAGQA~\cite{wang2025videollm} & 5.07 & 0.00 & 5.07 \\
DiDeMo~\cite{anne2017localizing} & 4.80 & 0.00 & 4.80 \\
VideoChat3~\cite{li2026videochat3fullyopenvideo} & 4.25 & 0.17 & 4.42 \\
PerceptionTest~\cite{patraucean2023perception} & 0.93 & 2.69 & 3.62 \\
UVQA~\cite{yoon2025can} & 0.00 & 3.58 & 3.58 \\
COIN~\cite{tang2019coin} & 0.30 & 2.54 & 2.84 \\
VALOR32K-AVQA~\cite{riahi2025valor32k} & 0.00 & 2.63 & 2.63 \\
QuerYD~\cite{oncescu2021queryd} & 0.53 & 1.74 & 2.26 \\
AVA~\cite{gu2018ava} & 0.00 & 1.90 & 1.90 \\
QVHighlights~\cite{lei2021detecting} & 1.76 & 0.00 & 1.76 \\
CIRCLE-40K~\cite{yang2025cambrian} & 0.00 & 1.69 & 1.69 \\
Ego4D~\cite{grauman2022ego4d} & 0.00 & 1.61 & 1.61 \\
Charades~\cite{sigurdsson2016hollywood} & 0.00 & 1.57 & 1.57 \\
ActivityNet~\cite{krishna2017dense} & 1.17 & 0.33 & 1.50 \\
ActivityNet-QA~\cite{yu2019activitynet} & 0.00 & 1.29 & 1.29 \\
NExT-QA~\cite{xiao2021next} & 0.00 & 1.27 & 1.27 \\
TACoS~\cite{regneri2013grounding} & 0.97 & 0.00 & 0.97 \\
YouTube-Collected~\cite{zeng2026video} & 0.65 & 0.00 & 0.65 \\
Social-IQ~\cite{zadeh2019social} & 0.00 & 0.52 & 0.52 \\
OOPS~\cite{epstein2020oops}& 0.43 & 0.00 & 0.43 \\
CounterVid~\cite{poppi2026countervid} & 0.00 & 0.29 & 0.29 \\
FunQA~\cite{xie2024funqa} & 0.23 & 0.00 & 0.23 \\
CogStream~\cite{zhao2026cogstream} & 0.00 & 0.19 & 0.19 \\
VST-Training-Data~\cite{guan2026video} & 0.00 & 0.13 & 0.13 \\
Countix~\cite{zhang2025period} & 0.00 & 0.12 & 0.12 \\
YouCook2~\cite{zhou2018towards} & 0.07 & 0.00 & 0.07 \\
OnlineIT-Drive~\cite{zeng2026streamforest} & 0.00 & 0.02 & 0.02 \\
JoyVL~\cite{yao2026joyaivlinteractionrealtimevisionlanguageinteraction} & $<0.01$ & 0.00 & $<0.01$ \\
\midrule
\textbf{Total} & \textbf{36.66} & \textbf{63.34} & \textbf{100.00} \\
\bottomrule
\end{tabularx}
\end{table}

% -------------------------------------------------------------------------
\subsection{Training Configuration}
\label{sec:supp_config}

\noindent \textbf{Scale-specific optimization.}
Table~\ref{tab:supp_train_config} summarizes the training and memory hyperparameters for StreamDLM-4B and StreamDLM-8B, initialized from Bard-VL-4B and Bard-VL-8B, respectively. The four KV cache caps sum to 27,648 token positions, leaving 5,120 within the 32,768-token context limit. Both models train on $32$ NVIDIA H100 SXM 80GB GPUs across four nodes ($1$ session per GPU, AdamW, BF16 forward autocast, fixed random seed). StreamDLM-4B performs full-parameter fine-tuning of the language backbone in BF16 under ZeRO-2. StreamDLM-8B attaches LoRA adapters (rank $128$, scaling $\alpha = 256$, dropout $0.05$) to all attention and MLP linear projections, using FP32 for trainable weights under DDP and applying weight decay ($0.01$) only to visual merger matrices.

\begin{table}[!t]
\centering
\caption{Training configurations and context budgets.}
\label{tab:supp_train_config}
\label{tab:training_setup}
\suppSetupTableStyle
\setlength{\tabcolsep}{2pt}
\begin{tabular*}{\linewidth}{@{\extracolsep{\fill}}lrr@{}}
\toprule
\textbf{Hyperparameter} & \textbf{StreamDLM-4B} & \textbf{StreamDLM-8B} \\
\midrule
Language-model tuning & Full & LoRA \\
LoRA rank / $\alpha$ / dropout & --- & $128/256/0.05$ \\
Language / LoRA LR & $1\times10^{-5}$ & $5\times10^{-5}$ \\
Visual merger LR & $5\times10^{-6}$ & $1\times10^{-5}$ \\
Control-token LR & $5\times10^{-5}$ & $1\times10^{-4}$ \\
Trainable weight dtype & BF16 & FP32 \\
Distributed strategy & ZeRO-2 & DDP \\
GPUs / batch (sessions) & $32/32$ & $32/32$ \\
Recent window / KV cap & $80\,\mathrm{s}/12{,}288$ & $80\,\mathrm{s}/12{,}288$ \\
Long-term visual KV cap & $12{,}288$ & $12{,}288$ \\
Text KV caps ($\mathcal H_o/\mathcal H_a$) & $1{,}024/2{,}048$ & $1{,}024/2{,}048$ \\
Segment limits $(E_{\max},W_{\max})$ & $(16,1.2{\times}10^7)$ & $(16,1.2{\times}10^7)$ \\
Training steps & 18,700 & 18,700 \\
Cumulative training time & $\approx 140\,\mathrm{h}$ & $\approx 100\,\mathrm{h}$ \\
\bottomrule
\end{tabular*}
\end{table}

% =========================================================================
\clubpenalty=150 \widowpenalty=150 \displaywidowpenalty=50
\section{Evaluation and Efficiency Protocols}
\label{sec:supp_eval}

This section details the online visibility rules, benchmark evaluation protocols across all five streaming benchmarks (seven evaluation tasks), and the single-GPU efficiency benchmark (\S4.4).

% -------------------------------------------------------------------------
\subsection{Online Visibility and Baseline Protocols}
\label{sec:supp_eval_input}

At a query time $t$, both online and offline models can use only video observed through $t$. For ordinary question answering, the benchmark supplies the query time. Online models process frames as they arrive and answer from the history they have retained. For each question, offline models typically receive up to 32 frames sampled uniformly from the video available at the query time; the exact budget depends on the model and task.

For proactive evaluation, the model must also determine \emph{when} to answer. Online models use their native response controllers where available. StreamDLM receives the question once, ingests two frames per second in one-second intervals, and checks the gate once per second of video time. Its ordinary QA runs use the same default input cadence and bounded history (\S3.3). For offline models, we make a fresh query at each polling time using up to 32 frames sampled uniformly from the video visible then. On StreamingBench-PO, we poll once per second and score the first triggered response. For OVO-Timing, we poll at the specified OVO-Bench times and use Em-Garde to score the timing of events inferred from successive answers.

We use the highest published score when its protocol and metric match; otherwise we evaluate locally, prioritizing official inference code. For MOSS-VL-Realtime, our reproduction runs yielded scores that differed substantially from those reported in the original paper. We therefore use our local evaluation results for this model across all benchmarks. Online baselines retain their native frame rates (typically 1 or 2\,FPS), frame grouping, memory, and decoding. The unified RTV-Bench reevaluation, including StreamDLM, uses 1\,FPS for visible intervals up to 180\,s and 180 uniformly spaced frames over the full visible interval for longer inputs. Published results retain their original settings.

% -------------------------------------------------------------------------
\subsection{Benchmark Scoring and Aggregation}
\label{sec:supp_eval_metrics}

All evaluations follow the official scoring scripts and metric definitions of each benchmark, summarized in Table~\ref{tab:supp_eval_protocols}. One practical distinction to note is macro-averaging versus question-level pooling on OVO-Bench and OVBench: because subtask sizes differ, identical predictions yield $68.68\%$ and $65.99\%$ with the official macro-averaging used in Table~1 of the main paper versus $65.17\%$ and $62.62\%$ under question-level pooling for StreamDLM-8B. On proactive tasks, StreamingBench-PO requires the first non-empty output to match the reference text within a $\pm 2\text{ s}$ window, while OVO-Timing performs greedy one-to-one matching of all non-empty outputs within $\pm 2\text{ s}$ and averages task-level F1 across CRR, REC, and SSR.

\begin{table}[!htbp]
\centering
\caption{Benchmark-recommended metrics and aggregation rules for the seven evaluation tasks.}
\label{tab:supp_eval_protocols}
\suppSetupTableStyle
\setlength{\tabcolsep}{2.5pt}
\begin{tabularx}{\columnwidth}{@{}l>{\raggedright\arraybackslash}X@{}}
\toprule
\textbf{Task} & \textbf{Reported score} \\
\midrule
OVO-Bench & Mean subtask accuracy within each group, then mean over the three groups. \\
OVBench & Mean accuracy over 16 subtasks. \\
StreamingBench-RT & Accuracy pooled over questions. \\
StreamingBench-PO & First response contains the target text within $\pm2$\,s; mean over episodes. \\
SVCBench & GPA, MoC, UDA, each averaged over applicable trajectories. \\
RTV-Bench & Group-gated accuracy: credit requires correct prerequisite answers. \\
OVO-Timing & One-to-one matches within $\pm2$\,s; mean F1 over CRR, REC, and SSR. \\
\bottomrule
\end{tabularx}
\end{table}

\begin{table}[!t]
\centering\small\renewcommand{\arraystretch}{1.08}
\caption{Streaming benchmark results and inference speed. We evaluate OneStreamer locally on StreamingBench-PO and SVCBench, where the authors did not report results. For the other benchmarks, we use previously reported scores for OneStreamer. We measure each configuration's time to first token and generation rate on one NVIDIA A100-SXM4-80GB GPU.}
\label{tab:supp_concurrent}
\setlength{\tabcolsep}{2pt}
\begin{tabular*}{\columnwidth}{@{\extracolsep{\fill}}lrrr@{}}
\toprule
Metric & OneStreamer & StreamDLM & StreamDLM \\
       & (4B) & (4B) & (8B) \\
\midrule
OVO-Bench overall       & 72.1 & 64.36 & 68.68 \\
StreamingBench-RT       & 86.9 & 78.84 & 83.12 \\
OVBench average         & 66.8 & 60.83 & 65.99 \\
SVCBench GPA            & 32.22 & 28.25 & 32.33 \\
SVCBench MoC            & 69.79 & 71.28 & 83.53 \\
SVCBench UDA            & 43.85 & 37.66 & 44.59 \\
StreamingBench-PO       & 44.40 & 56.80 & 65.60 \\
OVO-Timing mean F1      & 41.6 & 35.35 & 36.14 \\
\midrule
Time to first token (s) & 18.6213 & 0.0975 & 0.1009 \\
Generation (tokens/s)   & 36.19 & 194.40 & 139.11 \\
\bottomrule
\end{tabular*}
\end{table}

\subsection{Inference Efficiency Protocol (\S4.4)}
\label{sec:supp_eval_eff}
\label{sec:efficiency_settings}

Following \S4.4, we use a single NVIDIA A100-SXM4-80GB GPU and run three trials for each OVBench video duration (32, 64, 128, and 256\,s). We allow up to 256 output tokens and retain each model's native stopping criteria. We exclude model loading, media preparation and decoding, and warm-up from timing. TTFT measures the elapsed time from request submission to the first non-empty output, including pending input processing. Peak GPU memory records the maximum allocated memory across video ingestion and generation, including resident KV caches. Both metrics are averaged across the three repetitions per duration and then equally across the four durations. Because block diffusion can emit multiple tokens in the first batch, TPS divides actual tokens emitted \emph{after} the first non-empty batch (including natural $\texttt{[EOS]}$ and excluding other control tokens) by the interval from the first to the last output event; token counts and intervals are summed over the three repetitions per duration before averaging rates across the four durations.

We measure system-level inference speed using each model's inference implementation under the hardware and workload settings described above. StreamDLM-4B and StreamDLM-8B use BF16 and four-token blocks with up to four denoising steps and early termination.

For the OneStreamer comparison (Table~\ref{tab:supp_concurrent}), we release frames in timestamp order at 2\,FPS with a budget of 131,072 pixels per frame and submit the question at the end of each video prefix. OneStreamer and StreamDLM-8B use greedy decoding and reproduce the same 115-word passage verbatim in all 12 trials, stopping at \texttt{[EOS]}. StreamDLM-8B uses a confidence threshold of 0.5 for unmasking.

% -------------------------------------------------------------------------
\begingroup
\setlength{\textfloatsep}{12pt plus 1pt minus 1pt}
\setlength{\intextsep}{11pt plus 1pt minus 1pt}
\section{Extended Evaluation and Analysis}
\label{sec:extended_analysis}

\subsection{Comparison with a Finetuned Autoregressive Baseline}
\label{sec:ar-baseline}

We fine-tune Qwen3-VL-8B-Instruct on the same streaming training-data pool as StreamDLM-8B and denote the resulting autoregressive baseline by Qwen3-VL$^{*}$. We match the LoRA hyperparameters, learning rates, global batch size, and training hardware. Both models use the same response-triggering settings for streaming video. Specifically, on proactive tasks (StreamingBench-PO and OVO-Timing), Qwen3-VL$^{*}$ receives the question once and checks its learned \texttt{WAIT}/\texttt{ANSWER} gate after each one-second video update. The gate is trained with the same class-balanced control loss via causal next-token prediction over \texttt{<wait>} and \texttt{<answer>}. Across updates, it maintains separate KV histories for observations and generated answers, retaining recent visual observations in an 80-second window and compressing older ones as needed. For answer generation, Qwen3-VL$^{*}$ retains its original architecture and is trained to predict the next token, whereas StreamDLM-8B is trained to denoise answer blocks. We evaluate both models on the same benchmark items using the scoring protocols from the main paper. We measure inference speed using the fixed-passage protocol from the main paper. Unlike the native Qwen3-VL-8B inference setup reported in Table~4 of the main paper, our Qwen3-VL$^{*}$ setup uses an optimized autoregressive decoding path and follows the same streaming KV-cache policy and 80-second visual window as StreamDLM-8B to align their streaming memory management.

\begin{table}[t]
    \centering\small\renewcommand{\arraystretch}{1.08}
    \caption{Streaming video quality and inference speed for the two 8B models. Quality scores compare StreamDLM with the fine-tuned Qwen3-VL$^{*}$ baseline; differences are in points. The final two rows report time to first token and generation throughput on the fixed passage.}
    \label{tab:ar-quality-speed}
    \setlength{\tabcolsep}{2pt}
    \begin{tabular*}{\columnwidth}{@{\extracolsep{\fill}}lrrr@{}}
        \toprule
        Metric                   & Qwen3-VL$^{*}$ & StreamDLM & Change \\
                                 & (8B)     & (8B)      &        \\
        \midrule
        OVO-Bench overall        & 66.84 & 68.68 & +1.84 \\
        OVBench average          & 62.11 & 65.99 & +3.88 \\
        StreamingBench-RT        & 82.88 & 83.12 & +0.24 \\
        StreamingBench-PO        & 60.40 & 65.60 & +5.20 \\
        RTV-Bench score          & 21.68 & 26.41 & +4.73 \\
        SVCBench GPA             & 31.88 & 32.33 & +0.45 \\
        SVCBench MoC             & 81.83 & 83.53 & +1.70 \\
        SVCBench UDA             & 44.25 & 44.59 & +0.34 \\
        OVO-Timing mean F1       & 36.10 & 36.14 & +0.04 \\
        \midrule
        Time to first token (ms) & 250.22 & 100.89 & $-59.7\%$     \\
        Generation (tokens/s)    & 50.00 & 139.11 & $2.78\times$ \\
        \bottomrule
    \end{tabular*}
\end{table}

StreamDLM-8B scores higher than Qwen3-VL$^{*}$ on all nine quality metrics (Table~\ref{tab:ar-quality-speed}). The largest gains are on StreamingBench-PO (+5.20 points), RTV-Bench (+4.73 points), and OVBench (+3.88 points).  On the fixed-passage benchmark, StreamDLM-8B generates 139.11 tokens/s versus 50.00 for Qwen3-VL$^{*}$ ($2.78\times$ the throughput). Time to first token is 100.89\,ms for StreamDLM-8B and 250.22\,ms for Qwen3-VL$^{*}$.

These results suggest that blockwise decoding can improve generation efficiency while preserving overall streaming video understanding performance and may even improve performance on some tasks. Qwen3-VL$^{*}$ generates one token at a time, whereas StreamDLM can accept multiple tokens in a single denoising step. This reduces the number of sequential steps needed to generate an answer and helps explain the higher generation throughput in Table~\ref{tab:ar-quality-speed}. A potential benefit for answer quality comes from the order in which tokens are resolved. Within a block, StreamDLM can leave uncertain positions masked while accepting more confident predictions elsewhere. At the next step, it predicts the remaining tokens using both the visual context and tokens already accepted to their left or right. This additional context may help the model resolve uncertain predictions more accurately.

\subsection{Comparison with Concurrent Work}
\label{sec:supp_concurrent}
OneStreamer~\cite{zeng2026onestreamer}, a concurrent work, builds a textual history as video frames arrive, generating local captions and summaries that remain available after the corresponding frames leave its visual window. Table~\ref{tab:supp_concurrent} compares StreamDLM with OneStreamer using previously reported scores where available, supplemented by our own evaluations. With our once-per-second response checks on StreamingBench-PO, OneStreamer scores 44.40\% versus 65.60\% for StreamDLM-8B. The largest SVCBench gap is in MoC: StreamDLM-8B scores 83.53 and OneStreamer 69.79.

To test whether OneStreamer can update its video history as fast as new frames arrive, we measure how long a query waits for pending updates (protocol in \S\ref{sec:supp_eval_eff}). OneStreamer's mean queueing delay grows from 0.13\,s on the 32-second stream to 54.18\,s on the 256-second stream. On the latter, the first token takes only 0.52\,s once generation starts, but reaches the user 54.96\,s after the query. When captions lag behind incoming frames, queued history updates can dominate latency despite fast decoding.

\subsection{Additional Results on Inference Efficiency}
\label{sec:supp_latency_denoising}
We supplement \S4.4 of the main paper with a latency breakdown and throughput measurements at different denoising step counts. StreamDLM-8B's mean TTFT is 100.89\,ms: 0.15\,ms in the question queue, 67.79\,ms in prefill, and 32.95\,ms from prefill to first output. Throughput counts tokens emitted after the first block, including \texttt{[EOS]} (\S\ref{sec:supp_eval_eff}).

Each trial produced 132 tokens in 33 blocks. Across 12 trials, 396 denoising forwards for 396 blocks give \textbf{1.00 step per block}. Successive commits averaged \textbf{28.84\,ms} apart (384 intervals; Table~\ref{tab:supp_latency_steps}), including denoising, KV commitment, and runtime overhead.

\begin{table}[!htbp]
\centering\small\renewcommand{\arraystretch}{1.08}
\caption{StreamDLM-8B latency by input duration, averaged over three trials.}
\label{tab:supp_latency_steps}
\setlength{\tabcolsep}{3pt}
\begin{tabular*}{\columnwidth}{@{\extracolsep{\fill}}rrr@{}}
\toprule
Video (s) & First token (ms) & Block interval (ms) \\
\midrule
32 & 72.51 & 26.42 \\
64 & 85.31 & 30.55 \\
128 & 115.95 & 28.65 \\
256 & 129.78 & 29.75 \\
\bottomrule
\end{tabular*}
\end{table}

Table~\ref{tab:supp_fixed_denoising} reports fixed-passage decoding with two to four denoising steps per block on a single NVIDIA A100-SXM4-80GB GPU. With adaptive decoding on all 3,035 OVO-Bench questions, StreamDLM-4B and StreamDLM-8B averaged 1.382 and 1.336 steps per block, respectively. For each model, we assign each block in the 115-word passage benchmark one, two, or three denoising steps to approximate its empirical step-count distribution, and execute all assigned steps. The resulting throughput is 186.20 and 133.22 tokens/s, with TTFTs of 92.37 and 105.73\,ms, respectively.

On a single NVIDIA A100-SXM4-80GB GPU, we additionally evaluate StreamDLM-8B on 250 clips from ActivityNet Captions, each 24 seconds long. The model receives one question at the start of each clip and decides when to respond as new frames arrive. Across the 250 clips, it produces 3,525 utterances with a generation throughput of 107.19 tokens/s. To measure throughput at a fixed output length, we also calibrate the 115-word passage benchmark to the denoising workload observed on these clips: an average of 2.04 denoising steps per four-token answer block. We run three trials at each of four video prefix durations (32, 64, 128, and 256 seconds). Weighting the four durations equally gives a mean throughput of 103.39 tokens/s.

\begin{table}[!htbp]
\caption{StreamDLM inference speed at fixed denoising steps.}
\label{tab:supp_fixed_denoising}
\centering\small\renewcommand{\arraystretch}{1.08}
\setlength{\tabcolsep}{2pt}
\begin{tabular*}{\linewidth}{@{\extracolsep{\fill}}lcrrr@{}}
\toprule
Model & Params & Steps & Tokens/s & TTFT (ms) \\
\midrule
StreamDLM & 4B & 2 & 150.09 & 98.19 \\
StreamDLM & 4B & 3 & 114.50 & 104.58 \\
StreamDLM & 4B & 4 & 92.01 & 115.81 \\
\midrule
StreamDLM & 8B & 2 & 104.93 & 115.34 \\
StreamDLM & 8B & 3 & 79.70 & 127.58 \\
StreamDLM & 8B & 4 & 63.73 & 141.96 \\
\bottomrule
\end{tabular*}
\end{table}

\subsection{Additional Ablation Studies}
\setlength{\intextsep}{11pt plus 1pt minus 1pt}

\begin{table}[!htbp]
    \centering
    \caption{Pre-answer gating ablations for StreamDLM-8B on StreamingBench-PO and OVO-Bench. All variants are evaluated after 4,000 training steps.}
    \label{tab:speech-gate-compact}
    \begingroup
    \small\renewcommand{\arraystretch}{1.08}
    \setlength{\tabcolsep}{2pt}
    \begin{tabular*}{\columnwidth}{@{\extracolsep{\fill}}lrr@{}}
        \toprule
        Gate & StreamingBench-PO & OVO-Bench \\
        \midrule
        StreamDLM             & \textbf{48.40} & \textbf{61.61} \\
        Allow access     & 17.60 & 59.41 \\
        Block access     & 38.40 & 58.42 \\
        Count-weighted loss & 37.60 & 60.76 \\
        No class balance &  0.40 & 61.42 \\
        \bottomrule
    \end{tabular*}
    \endgroup
\end{table}

\noindent \textbf{Pre-answer gating.}
The response-control ablation in the main paper compares StreamDLM's pre-answer gate with \emph{Joint Decoding}, which places control tokens inside the answer block and trains them under the same token loss without gate-class balancing. Here we test \emph{answer visibility at the gate} (Table~\ref{tab:speech-gate-compact}). Both access variants evaluate a separate gate query alongside the first four-token answer block, retain separate gate and text loss normalization, and balance \texttt{WAIT} and \texttt{ANSWER}. They differ only in whether the gate can read the answer tokens. Suppose ``door'' is left unmasked in the reference answer. Both variants can use ``door'' to predict the missing answer tokens. Under \emph{Allow access}, the gate can also read ``door'' when choosing \texttt{WAIT} or \texttt{ANSWER}. Under \emph{Block access}, the gate reads only the observed context, not the answer block.

\begin{table}[!t]
\centering\captionsetup{position=top,skip=4pt}
\setlength{\tabcolsep}{2pt}
\caption{Training speed on eight H100 GPUs. \emph{w/o} denotes the component removed from StreamDLM. Speedup is the throughput of StreamDLM divided by that of the row.}
\label{tab:h100-training-speed}
\suppResultsTableStyle
\setlength{\aboverulesep}{0.25ex}
\setlength{\belowrulesep}{0.25ex}
\begin{tabularx}{\columnwidth}{@{}>{\raggedright\arraybackslash}Xrrr@{}}
\toprule
Configuration & s/step & Sessions/s & Speedup \\
\midrule
StreamDLM & \textbf{53.08} & \textbf{0.603} & $1.00\times$ \\
w/o Source packing           & 77.89  & 0.411 & $1.47\times$ \\
w/o Query packing            & 133.90 & 0.239 & $2.52\times$ \\
w/o Clean KV packing         & 111.17 & 0.288 & $2.09\times$ \\
w/o Bounded backpropagation  & 108.97 & 0.294 & $2.05\times$ \\
\bottomrule
\end{tabularx}
\end{table}

The two loss variants keep the original StreamDLM gate, which decides whether to speak before any answer tokens are decoded. \emph{Count-weighted loss} preserves class balance but weights the gate and text losses by their respective numbers of supervised decisions and text predictions. \emph{No class balance} retains separate normalization and weights actions by frequency.

StreamDLM scores highest on OVO-Bench (61.61\%). The variants differ most on proactive answering, where StreamDLM reaches 48.40\% on StreamingBench-PO. \emph{Allow access} trails by 30.80 points: during training, visible reference words may give the gate a cue to speak that is absent at inference. \emph{Block access} trails by 10.00 points: even without that cue, predicting the first answer block before the decision to speak is settled may make the opening response harder to learn. \emph{Count-weighted loss} trails by 10.80 points: when text predictions outnumber gate decisions, the gate receives less relative weight. \emph{No class balance} trails by 48.00 points: \texttt{WAIT} labels outnumber \texttt{ANSWER} labels in training, so without reweighting the gate may learn to keep waiting even when it should answer.

In the response-control ablation in the main paper, unmodified Bard-VL-8B streams two frames per second into persistent memory. It answers OVO-Bench queries from memory. On proactive tasks, we provide the question once and invoke the decoder every second. In Table~1 of the main paper, each offline query instead receives up to 32 uniformly sampled frames from the visible prefix. StreamingBench-PO polls every second until a positive check triggers an answer.

\noindent \textbf{Shared context packing.}
We benchmark LoRA training of StreamDLM-8B on a single node with eight H100 GPUs, ablating one component from \S\ref{sec:supp_packing} at a time (Table~\ref{tab:h100-training-speed}). After eight warm-up steps, StreamDLM and \emph{w/o Source packing} processed all 64 measured batches in full. Without query packing, clean KV packing, or bounded backpropagation, GPU out-of-memory errors caused 85, 48, and 12 sessions to be skipped, respectively. To compare identical workloads, we report the 18 batches for which all five configurations processed all 32 sessions (576 sessions total). The main paper compares only StreamDLM and \emph{w/o Query packing}, using all 21 batches completed by both.

StreamDLM processes 0.603 sessions/s at 53.08\,s per step. Removing query packing has the largest effect: 133.90\,s per step and 0.239 sessions/s. On the same batches, StreamDLM is 2.09$\times$ faster than without clean KV packing and 2.05$\times$ faster than without bounded backpropagation.

\begin{table}[!t]
\centering\captionsetup{position=top,skip=4pt}
\caption{Mean judge scores on 150 ActivityNet Captions clips (1--10 scale).}
\label{tab:supp_refresh_scores}
\suppResultsTableStyle
\setlength{\aboverulesep}{0.25ex}
\setlength{\belowrulesep}{0.25ex}
\begin{tabular*}{\linewidth}{@{\extracolsep{\fill}}lr@{}}
\toprule
Configuration & Mean score \\
\midrule
StreamDLM (block refresh) & \textbf{6.23} \\
Frozen observation history & 6.00 \\
Step refresh & 5.91 \\
Frozen answer history & 4.48 \\
Answer KV in observation history & 5.67 \\
\bottomrule
\end{tabular*}
\end{table}

\noindent \textbf{Blockwise context management.}
We randomly sampled 150 clips from ActivityNet Captions for evaluation, excluding duplicate clips and any clips used in StreamDLM training. Each 24-second clip contains annotated event onsets in at least two distinct 6-second intervals. A description can therefore span multiple answer blocks, and the model must observe new frames to describe later events. The model receives the prompt (Box~\ref{box:generation}) only with the first frame. As frames arrive at 4\,FPS, the model describes the opening action and decides when to speak again. For each clip, GPT-6.2 Astra~\cite{openai2026gpt62astra} gave each of five anonymized responses one score for accuracy and timing (1--10; Box~\ref{box:judge}).

All variants use the same 8B checkpoint, four-token blocks, and four denoising steps per block. StreamDLM holds the visual context fixed while denoising each answer block. Frames arriving during that block become available at the start of the next block in the same utterance. With \emph{Frozen observation history}, all blocks use the visual context available at the start of the utterance. After \texttt{[EOS]}, the model can use the new frames to decide whether to speak again in response to the same question. In both configurations, each completed block is added to the answer history. \emph{Step refresh} updates the visual context before every denoising step. Two controls retain block-level refresh: \emph{Frozen answer history} hides earlier blocks of the current utterance from later blocks; \emph{Answer KV in observation history} stores answer KV with visual observations in a shared history.

StreamDLM averages 6.23 out of 10 (Table~\ref{tab:supp_refresh_scores}). If an action changes while a block is being decoded, StreamDLM can describe it in the next block of the same utterance; \emph{Frozen observation history} (6.00) must wait until the next utterance. \emph{Step refresh} (5.91) can use new frames sooner, but its denoising steps need not see the same action while refining a single block. Under block-level refresh, \emph{Frozen answer history} (4.48) cannot read what it has already said in the current utterance. \emph{Answer KV in observation history} scores 5.67 with shared KV versus 6.23 with separate answer KV.

\par\endgroup
\setlength{\textfloatsep}{8pt plus 1pt minus 1pt}
\clubpenalty=10000 \widowpenalty=10000 \displaywidowpenalty=10000
\newcounter{ACstreamPrompt}
\begingroup
\vspace{8pt}
\setlength{\fboxsep}{5pt}
\noindent\fcolorbox[HTML]{A6CDE3}{EAF5FC}{%
  \begin{minipage}{\dimexpr\linewidth-2\fboxsep-2\fboxrule\relax}
  \refstepcounter{ACstreamPrompt}\label{box:generation}%
  \linespread{1}\fontsize{9}{10.45}\selectfont\raggedright\textbf{Box~\theACstreamPrompt. Generation prompt}\par\nobreak
  How does the action change from one moment to the next? Describe the first view. On later views, tell me the concrete difference from the last thing you described: someone was doing one thing, then you see them do something else. If it is a different shot, introduce it as a new scene. Only describe visible changes.

Example: ``At first the woman is tying her shoes. Now she has stood up and is walking toward the door.'' Another example: ``The camera cuts from the orchestra to a couple dancing.'' Use those as examples of form, not of what this video contains.
  \end{minipage}}\par
\vspace{7pt}
\noindent\fcolorbox[HTML]{A6CDE3}{EAF5FC}{%
  \begin{minipage}{\dimexpr\linewidth-2\fboxsep-2\fboxrule\relax}
  \refstepcounter{ACstreamPrompt}\label{box:judge}%
  \linespread{1}\fontsize{9}{10.45}\selectfont\raggedright\textbf{Box~\theACstreamPrompt. Judge prompt}\par\nobreak
  Judge the five anonymous responses to each video independently on an overall 1-to-10 scale (10 is best). Inspect the timestamp-labelled video panels in order and the timestamped new assertions. Reward accurate description of the opening action and later visible changes at the right time. Do not invent meaning for an unintelligible fragment. Output one score for each of A through E for each video.
  \end{minipage}}\par
\vspace{7pt}
\endgroup
\clubpenalty=150 \widowpenalty=150 \displaywidowpenalty=50
\subsection{Detailed Experimental Results}
\label{sec:detailed_results}

Tables~\ref{tab:detailed_ovo}--\ref{tab:detailed_timing} provide detailed comparisons of StreamDLM-4B and StreamDLM-8B with AURA, Foresight, JoyAI-VL-Interaction, StreamForest, and VideoChat3. The results cover real-time visual perception, backward tracing, and forward active responding on OVO-Bench; the subtasks of OVBench; the real-time tasks of StreamingBench; the evaluation dimensions of RTV-Bench and SVCBench; and CRR, REC, and SSR on OVO-Timing. All scores are reported on a 0--100 scale, with higher values indicating better performance. Boldface indicates the best result in each column, and the final column reports the overall score.

\section{Limitations}
\looseness=-1 StreamDLM allocates a large KV cache to visual history, which increases memory use. Under the inference protocol in \S4.4 of the main paper, StreamDLM-8B uses 43.5\,GB of peak GPU memory on average. Even with this budget, historical recall is limited: once frames leave the 80-second window, their visual representations may need to be compressed to fit in long-term memory. A later question may therefore refer to a brief action or visual detail that was not retained. StreamDLM also processes only video frames and text, so it cannot use audio cues such as speech or off-screen sounds.

\par
\onecolumn\raggedbottom
\begingroup
\noindent\begin{minipage}{\textwidth}\captionsetup{type=table,position=top,skip=4pt}
\centering
\caption{OVO-Bench accuracy. Avg. averages subtasks within a group; Overall averages the three groups.}
\label{tab:detailed_ovo}
\begingroup
\suppResultsTableStyle
\setlength{\tabcolsep}{1.8pt}
\setlength{\aboverulesep}{0.25ex}
\setlength{\belowrulesep}{0.25ex}
\begin{tabularx}{\textwidth}{@{}>{\raggedright\arraybackslash}p{0.22\textwidth}>{\centering\arraybackslash}p{0.065\textwidth}*{7}{>{\centering\arraybackslash}X}@{}}
\toprule
\textbf{Model} & \textbf{Params} & \multicolumn{7}{c}{\textbf{Realtime}} \\
 &  & OCR & ACR & ATR & STU & FPD & OJR & Avg. \\
\midrule
AURA & 8B & \textbf{89.90} & \textbf{79.80} & \textbf{80.20} & \textbf{70.20} & 77.20 & \textbf{81.50} & \textbf{79.80} \\
Foresight & 8B & 85.91 & 71.27 & 75.88 & 64.97 & 78.81 & 73.25 & 75.02 \\
JoyAI-VL-Interaction & 8B & 80.50 & 60.60 & 77.60 & 65.20 & \textbf{79.20} & 62.00 & 70.90 \\
StreamForest & 7B & 68.46 & 53.21 & 71.55 & 47.75 & 65.35 & 60.87 & 61.20 \\
VideoChat3 & 4B & 85.90 & 71.60 & 79.30 & 55.10 & 73.30 & 71.20 & 72.70 \\
\rowcolor{DRLight}
StreamDLM & 4B & 77.18 & 64.22 & 63.79 & 54.49 & 70.30 & 69.02 & 66.50 \\
\rowcolor{DRLight}
StreamDLM & 8B & 82.55 & 72.48 & 70.69 & 55.62 & 76.24 & 69.02 & 71.10 \\
\bottomrule
\end{tabularx}
\endgroup
\par\vspace{\suppPanelGap}
\begingroup
\suppResultsTableStyle
\setlength{\tabcolsep}{1.8pt}
\setlength{\aboverulesep}{0.25ex}
\setlength{\belowrulesep}{0.25ex}
\begin{tabularx}{\textwidth}{@{}>{\raggedright\arraybackslash}p{0.22\textwidth}>{\centering\arraybackslash}p{0.065\textwidth}*{9}{>{\centering\arraybackslash}X}@{}}
\toprule
\textbf{Model} & \textbf{Params} & \multicolumn{4}{c}{\textbf{Backward}} & \multicolumn{4}{c}{\textbf{Forward}} & \multicolumn{1}{c}{\textbf{Overall}} \\
 &  & EPM & ASI & HLD & Avg. & REC & SSR & CRR & Avg. &  \\
\midrule
AURA & 8B & 54.90 & 67.60 & 58.60 & 60.40 & 30.40 & 75.80 & \textbf{61.30} & 55.80 & 65.30 \\
Foresight & 8B & \textbf{62.20} & \textbf{69.50} & 33.60 & 55.10 & 32.10 & 77.20 & 59.80 & 56.37 & 62.16 \\
JoyAI-VL-Interaction & 8B & 58.90 & 62.80 & 23.10 & 48.30 & \textbf{45.80} & 70.30 & 58.80 & 58.30 & 59.10 \\
StreamForest & 7B & 58.92 & 64.86 & 32.26 & 52.02 & 32.81 & 70.59 & 57.08 & 53.49 & 55.57 \\
VideoChat3 & 4B & 56.90 & 58.10 & 47.30 & 54.10 & 23.80 & 67.20 & 55.00 & 48.70 & 58.50 \\
\rowcolor{DRLight}
StreamDLM & 4B & 55.89 & 60.14 & \textbf{100.00} & 72.01 & 33.95 & 78.06 & 51.67 & 54.56 & 64.36 \\
\rowcolor{DRLight}
StreamDLM & 8B & 60.27 & 68.92 & 93.55 & \textbf{74.25} & 41.55 & \textbf{79.33} & 61.25 & \textbf{60.71} & \textbf{68.68} \\
\bottomrule
\end{tabularx}
\endgroup
\end{minipage}
\par\vspace{\suppResultsGap}
\noindent\begin{minipage}{\textwidth}\captionsetup{type=table,position=top,skip=4pt}
\centering
\caption{StreamingBench real-time accuracy. Overall is accuracy pooled over questions.}
\label{tab:detailed_streaming_rt}
\begingroup
\suppResultsTableStyle
\setlength{\tabcolsep}{1.8pt}
\setlength{\aboverulesep}{0.25ex}
\setlength{\belowrulesep}{0.25ex}
\begin{tabularx}{\textwidth}{@{}>{\raggedright\arraybackslash}p{0.22\textwidth}>{\centering\arraybackslash}p{0.065\textwidth}*{10}{>{\centering\arraybackslash}X}>{\centering\arraybackslash}p{0.065\textwidth}@{}}
\toprule
\textbf{Model} & \textbf{Params} & OP & CR & CS & ATP & EU & TR & PR & SU & ACP & CT & Overall \\
\midrule
AURA & 8B & \textbf{87.50} & \textbf{84.40} & 93.40 & \textbf{89.50} & 83.20 & 87.90 & 86.10 & 76.40 & 82.40 & 47.70 & \textbf{83.20} \\
Foresight & 8B & 83.70 & 76.80 & 90.50 & 88.60 & 79.10 & 85.00 & 84.30 & 75.90 & 75.50 & 43.20 & 79.30 \\
JoyAI-VL-Interaction & 8B & 87.20 & 83.60 & 92.10 & 88.90 & 81.90 & \textbf{91.30} & \textbf{88.00} & 74.00 & 81.30 & 46.10 & 82.70 \\
StreamForest & 7B & 83.11 & 82.81 & 82.65 & 84.26 & 77.50 & 78.19 & 76.85 & 69.11 & 75.64 & \textbf{54.40} & 77.26 \\
VideoChat3 & 4B & 86.40 & \textbf{84.40} & \textbf{93.70} & 88.50 & 81.30 & 83.80 & 78.70 & \textbf{78.90} & \textbf{83.60} & 42.00 & 81.90 \\
\rowcolor{DRLight}
StreamDLM & 4B & 85.83 & 68.75 & 87.07 & 87.58 & 79.50 & 85.05 & 70.37 & 73.98 & 75.92 & 50.26 & 78.84 \\
\rowcolor{DRLight}
StreamDLM & 8B & 87.19 & 82.03 & 92.43 & 88.56 & \textbf{84.47} & 91.28 & 84.26 & 73.58 & 81.30 & 52.33 & 83.12 \\
\bottomrule
\end{tabularx}
\endgroup
\end{minipage}
\par\vspace{\suppResultsGap}
\noindent\begin{minipage}{\textwidth}\captionsetup{type=table,position=top,skip=4pt}
\centering
\caption{OVBench accuracy. Overall averages the 16 subtasks equally.}
\label{tab:detailed_ovbench}
\begingroup
\suppResultsTableStyle
\setlength{\tabcolsep}{1.8pt}
\setlength{\aboverulesep}{0.25ex}
\setlength{\belowrulesep}{0.25ex}
\begin{tabularx}{\textwidth}{@{}>{\raggedright\arraybackslash}p{0.22\textwidth}>{\centering\arraybackslash}p{0.065\textwidth}*{8}{>{\centering\arraybackslash}X}@{}}
\toprule
\textbf{Model} & \textbf{Params} & \multicolumn{3}{c}{\textbf{Future Prediction}} & \multicolumn{3}{c}{\textbf{Past Memory}} & \multicolumn{2}{c}{\textbf{Spatial Perception}} \\
 &  & AA & GSP & MP & AR & PR & TR & AL & OP \\
\midrule
AURA & 8B & 67.90 & \textbf{76.90} & 26.20 & 30.30 & 71.80 & 65.90 & 39.70 & 57.20 \\
Foresight & 8B & 55.13 & 64.60 & 27.91 & 33.75 & 67.01 & 61.96 & 33.44 & 42.72 \\
JoyAI-VL-Interaction & 8B & 55.10 & 54.00 & 29.80 & 50.00 & \textbf{73.60} & 87.10 & 35.00 & \textbf{89.50} \\
StreamForest & 7B & \textbf{69.20} & 60.00 & \textbf{34.40} & \textbf{50.90} & 64.90 & 82.20 & \textbf{56.60} & 87.90 \\
VideoChat3 & 4B & 59.00 & 62.00 & 23.30 & 46.30 & 68.70 & 79.60 & 35.60 & 85.90 \\
\rowcolor{DRLight}
StreamDLM & 4B & 51.28 & 71.09 & 23.31 & 31.56 & 59.61 & 86.50 & 30.00 & 78.33 \\
\rowcolor{DRLight}
StreamDLM & 8B & 58.97 & 76.11 & 20.86 & 45.00 & 70.38 & \textbf{88.34} & 33.13 & 87.93 \\
\bottomrule
\end{tabularx}
\endgroup
\par\vspace{\suppPanelGap}
\begingroup
\suppResultsTableStyle
\setlength{\tabcolsep}{1.8pt}
\setlength{\aboverulesep}{0.25ex}
\setlength{\belowrulesep}{0.25ex}
\begin{tabularx}{\textwidth}{@{}>{\raggedright\arraybackslash}p{0.22\textwidth}>{\centering\arraybackslash}p{0.065\textwidth}*{9}{>{\centering\arraybackslash}X}@{}}
\toprule
\textbf{Model} & \textbf{Params} & \multicolumn{2}{c}{\textbf{Spatiotemporal}} & \multicolumn{3}{c}{\textbf{Hallucination Verification}} & \multicolumn{3}{c}{\textbf{Temporal Perception}} & \multicolumn{1}{c}{\textbf{Overall}} \\
 &  & AT & OT & AP & OP & SV & AS & OES & SL &  \\
\midrule
AURA & 8B & 93.90 & 58.80 & 50.20 & 72.70 & 58.00 & 76.30 & 61.20 & 26.40 & 58.30 \\
Foresight & 8B & 72.61 & 47.52 & 62.23 & 72.00 & 55.22 & 78.60 & 44.62 & 26.78 & 52.88 \\
JoyAI-VL-Interaction & 8B & 94.80 & 56.50 & 48.10 & 76.60 & 51.30 & 73.00 & 90.20 & \textbf{32.30} & 62.30 \\
StreamForest & 7B & 95.20 & 61.20 & \textbf{69.10} & 72.90 & 54.00 & 64.20 & \textbf{92.60} & 30.60 & 65.37 \\
VideoChat3 & 4B & 95.20 & 58.10 & 63.90 & 69.40 & \textbf{58.20} & 81.90 & 85.60 & 26.50 & 62.50 \\
\rowcolor{DRLight}
StreamDLM & 4B & \textbf{99.57} & 66.46 & 60.94 & 63.38 & 56.06 & 85.58 & 84.62 & 25.05 & 60.83 \\
\rowcolor{DRLight}
StreamDLM & 8B & 98.26 & \textbf{66.77} & 66.52 & \textbf{77.54} & 57.65 & \textbf{86.98} & 90.46 & 30.99 & \textbf{65.99} \\
\bottomrule
\end{tabularx}
\endgroup
\end{minipage}
\par\vspace{\suppResultsGap}
\noindent\begin{minipage}{\textwidth}\captionsetup{type=table,position=top,skip=4pt}
\centering
\caption{RTV-Bench group-gated Score by capability. Overall uses the official aggregation.}
\label{tab:detailed_rtv}
\begingroup
\suppResultsTableStyle
\setlength{\tabcolsep}{1.8pt}
\setlength{\aboverulesep}{0.25ex}
\setlength{\belowrulesep}{0.25ex}
\begin{tabularx}{\textwidth}{@{}>{\raggedright\arraybackslash}p{0.22\textwidth}>{\centering\arraybackslash}p{0.065\textwidth}*{9}{>{\centering\arraybackslash}X}@{}}
\toprule
\textbf{Model} & \textbf{Params} & TP & VP & SP & IA & PU & GU & SR & FP & Overall \\
\midrule
AURA & 8B & 23.03 & 27.09 & 24.22 & 28.20 & 30.00 & \textbf{19.86} & 17.47 & \textbf{28.44} & 24.84 \\
Foresight & 8B & 20.61 & 18.73 & 14.88 & 20.00 & 19.33 & 14.44 & 12.65 & 18.75 & 17.50 \\
JoyAI-VL-Interaction & 8B & 22.12 & 26.58 & 16.61 & 21.31 & 25.33 & 16.61 & 13.86 & 20.31 & 20.57 \\
StreamForest & 7B & 23.64 & 23.29 & 23.53 & \textbf{29.18} & 23.00 & 14.44 & 14.46 & 25.62 & 22.21 \\
VideoChat3 & 4B & 13.03 & 17.47 & 21.11 & 20.98 & 14.67 & 10.11 & 12.35 & 22.81 & 16.60 \\
\rowcolor{DRLight}
StreamDLM & 4B & 21.52 & 23.29 & 19.03 & 22.95 & 21.38 & 14.08 & 14.16 & 24.69 & 20.30 \\
\rowcolor{DRLight}
StreamDLM & 8B & \textbf{26.06} & \textbf{33.67} & \textbf{27.34} & \textbf{29.18} & \textbf{30.67} & 16.61 & \textbf{19.28} & 26.25 & \textbf{26.41} \\
\bottomrule
\end{tabularx}
\endgroup
\end{minipage}
\par\vspace{\suppResultsGap}
\noindent\begin{minipage}{\textwidth}\captionsetup{type=table,position=top,skip=4pt}
\centering
\caption{SVCBench GPA by counting subtype. Overall uses the official aggregation.}
\label{tab:detailed_svc_gpa}
\begingroup
\suppResultsTableStyle
\setlength{\tabcolsep}{1.8pt}
\setlength{\aboverulesep}{0.25ex}
\setlength{\belowrulesep}{0.25ex}
\begin{tabularx}{\textwidth}{@{}>{\raggedright\arraybackslash}p{0.22\textwidth}>{\centering\arraybackslash}p{0.065\textwidth}*{9}{>{\centering\arraybackslash}X}@{}}
\toprule
\textbf{Model} & \textbf{Params} & \shortstack{E1\\Action} & \shortstack{E1\\Transit} & \shortstack{E2\\Episode} & \shortstack{E2\\Periodic} & \shortstack{O1\\Delta} & \shortstack{O1\\Snap} & \shortstack{O2\\Gain} & \shortstack{O2\\Unique} & Overall \\
\midrule
AURA & 8B & 21.93 & 28.59 & 26.44 & 0.72 & 38.78 & \textbf{32.00} & 47.44 & 31.38 & 28.91 \\
Foresight & 8B & 20.59 & 30.55 & 29.86 & 0.37 & 16.33 & 19.99 & 30.77 & 32.04 & 24.93 \\
JoyAI-VL-Interaction & 8B & 23.19 & 34.52 & \textbf{32.69} & 0.15 & 30.61 & 27.80 & \textbf{52.56} & \textbf{35.62} & 30.82 \\
StreamForest & 7B & 15.87 & 21.83 & 25.58 & 0.03 & 32.65 & 31.90 & 14.10 & 25.92 & 22.41 \\
VideoChat3 & 4B & 13.60 & 20.90 & 19.92 & 0.01 & 8.16 & 29.63 & 5.13 & 22.64 & 16.84 \\
\rowcolor{DRLight}
StreamDLM & 4B & 24.32 & 25.43 & 25.43 & \textbf{0.87} & 50.00 & 28.62 & 43.59 & 25.64 & 28.25 \\
\rowcolor{DRLight}
StreamDLM & 8B & \textbf{27.22} & \textbf{36.27} & 29.92 & 0.78 & \textbf{65.31} & 29.03 & \textbf{52.56} & 26.35 & \textbf{32.33} \\
\bottomrule
\end{tabularx}
\endgroup
\end{minipage}
\par\vspace{\suppResultsGap}
\noindent\begin{minipage}{\textwidth}\captionsetup{type=table,position=top,skip=4pt}
\centering
\caption{SVCBench MoC by counting subtype. Overall uses the official aggregation.}
\label{tab:detailed_svc_moc}
\begingroup
\suppResultsTableStyle
\setlength{\tabcolsep}{1.8pt}
\setlength{\aboverulesep}{0.25ex}
\setlength{\belowrulesep}{0.25ex}
\begin{tabularx}{\textwidth}{@{}>{\raggedright\arraybackslash}p{0.22\textwidth}>{\centering\arraybackslash}p{0.065\textwidth}*{6}{>{\centering\arraybackslash}X}@{}}
\toprule
\textbf{Model} & \textbf{Params} & \shortstack{E1\\Action} & \shortstack{E1\\Transit} & \shortstack{E2\\Episode} & \shortstack{E2\\Periodic} & \shortstack{O2\\Unique} & Overall \\
\midrule
AURA & 8B & 67.98 & 56.08 & 71.15 & 88.39 & 80.89 & 74.43 \\
Foresight & 8B & 82.18 & \textbf{85.94} & \textbf{83.28} & \textbf{88.84} & 85.77 & \textbf{84.55} \\
JoyAI-VL-Interaction & 8B & 77.57 & 75.17 & 78.05 & 74.55 & \textbf{91.30} & 82.66 \\
StreamForest & 7B & 80.91 & 80.21 & 80.98 & 83.04 & 86.61 & 83.27 \\
VideoChat3 & 4B & 61.64 & 50.69 & 64.08 & 61.61 & 63.91 & 62.29 \\
\rowcolor{DRLight}
StreamDLM & 4B & 74.78 & 63.54 & 74.94 & 19.02 & 76.63 & 71.28 \\
\rowcolor{DRLight}
StreamDLM & 8B & \textbf{84.22} & 79.69 & 82.41 & 61.76 & 88.10 & 83.53 \\
\bottomrule
\end{tabularx}
\endgroup
\end{minipage}
\par\vspace{\suppResultsGap}
\noindent\begin{minipage}{\textwidth}\captionsetup{type=table,position=top,skip=4pt}
\centering
\caption{SVCBench UDA by counting subtype. Overall uses the official aggregation.}
\label{tab:detailed_svc_uda}
\begingroup
\suppResultsTableStyle
\setlength{\tabcolsep}{1.8pt}
\setlength{\aboverulesep}{0.25ex}
\setlength{\belowrulesep}{0.25ex}
\begin{tabularx}{\textwidth}{@{}>{\raggedright\arraybackslash}p{0.22\textwidth}>{\centering\arraybackslash}p{0.065\textwidth}*{7}{>{\centering\arraybackslash}X}@{}}
\toprule
\textbf{Model} & \textbf{Params} & \shortstack{E1\\Action} & \shortstack{E1\\Transit} & \shortstack{E2\\Episode} & \shortstack{E2\\Periodic} & \shortstack{O1\\Snap} & \shortstack{O2\\Unique} & Overall \\
\midrule
AURA & 8B & 32.77 & 38.54 & 28.79 & 4.46 & 38.48 & 57.79 & 39.86 \\
Foresight & 8B & \textbf{42.41} & \textbf{63.54} & 40.98 & 27.68 & 26.95 & 60.61 & 47.28 \\
JoyAI-VL-Interaction & 8B & 40.10 & 50.52 & \textbf{46.09} & \textbf{46.43} & 32.92 & \textbf{63.63} & \textbf{49.80} \\
StreamForest & 7B & 28.29 & 39.76 & 31.67 & 13.39 & 34.57 & 51.52 & 37.36 \\
VideoChat3 & 4B & 28.50 & 25.35 & 30.86 & 14.73 & \textbf{40.12} & 45.47 & 34.93 \\
\rowcolor{DRLight}
StreamDLM & 4B & 28.70 & 34.38 & 27.18 & 5.98 & 30.25 & 56.79 & 37.66 \\
\rowcolor{DRLight}
StreamDLM & 8B & 41.90 & 54.51 & 36.03 & 22.06 & 26.95 & 58.02 & 44.59 \\
\bottomrule
\end{tabularx}
\endgroup
\end{minipage}
\par\vspace{\suppResultsGap}
\noindent\begin{minipage}{\textwidth}\captionsetup{type=table,position=top,skip=4pt}
\centering
\caption{OVO-Timing F1 at $\pm2$ seconds. Mean F1 averages CRR, REC, and SSR.}
\label{tab:detailed_timing}
\begingroup
\suppResultsTableStyle
\setlength{\tabcolsep}{2.6pt}
\setlength{\aboverulesep}{0.25ex}
\setlength{\belowrulesep}{0.25ex}
\begin{tabularx}{\linewidth}{@{}>{\raggedright\arraybackslash}p{0.22\linewidth}>{\centering\arraybackslash}X*{4}{>{\centering\arraybackslash}X}@{}}
\toprule
\textbf{Model} & \textbf{Params} & CRR & REC & SSR & Mean F1 \\
\midrule
AURA & 8B & 5.80 & 19.80 & 7.80 & 11.10 \\
Foresight & 8B & 18.97 & 27.92 & 14.90 & 20.60 \\
JoyAI-VL-Interaction & 8B & 14.60 & 36.20 & 9.30 & 20.00 \\
StreamForest & 7B & 2.08 & 29.67 & 10.11 & 13.95 \\
VideoChat3 & 4B & \textbf{31.40} & 45.80 & 23.50 & 33.60 \\
\rowcolor{DRLight}
StreamDLM & 4B & 25.66 & 52.17 & \textbf{28.21} & 35.35 \\
\rowcolor{DRLight}
StreamDLM & 8B & 19.65 & \textbf{62.06} & 26.71 & \textbf{36.14} \\
\bottomrule
\end{tabularx}
\endgroup
\end{minipage}
\par\endgroup

\clearpage
\flushbottom
\begingroup

\captionsetup{font=small,skip=5pt,justification=raggedright,singlelinecheck=false}

\begin{figure*}[p]
\section{Qualitative Cases}
\label{sec:qualitative_cases}
\centering
\includegraphics[width=\textwidth]{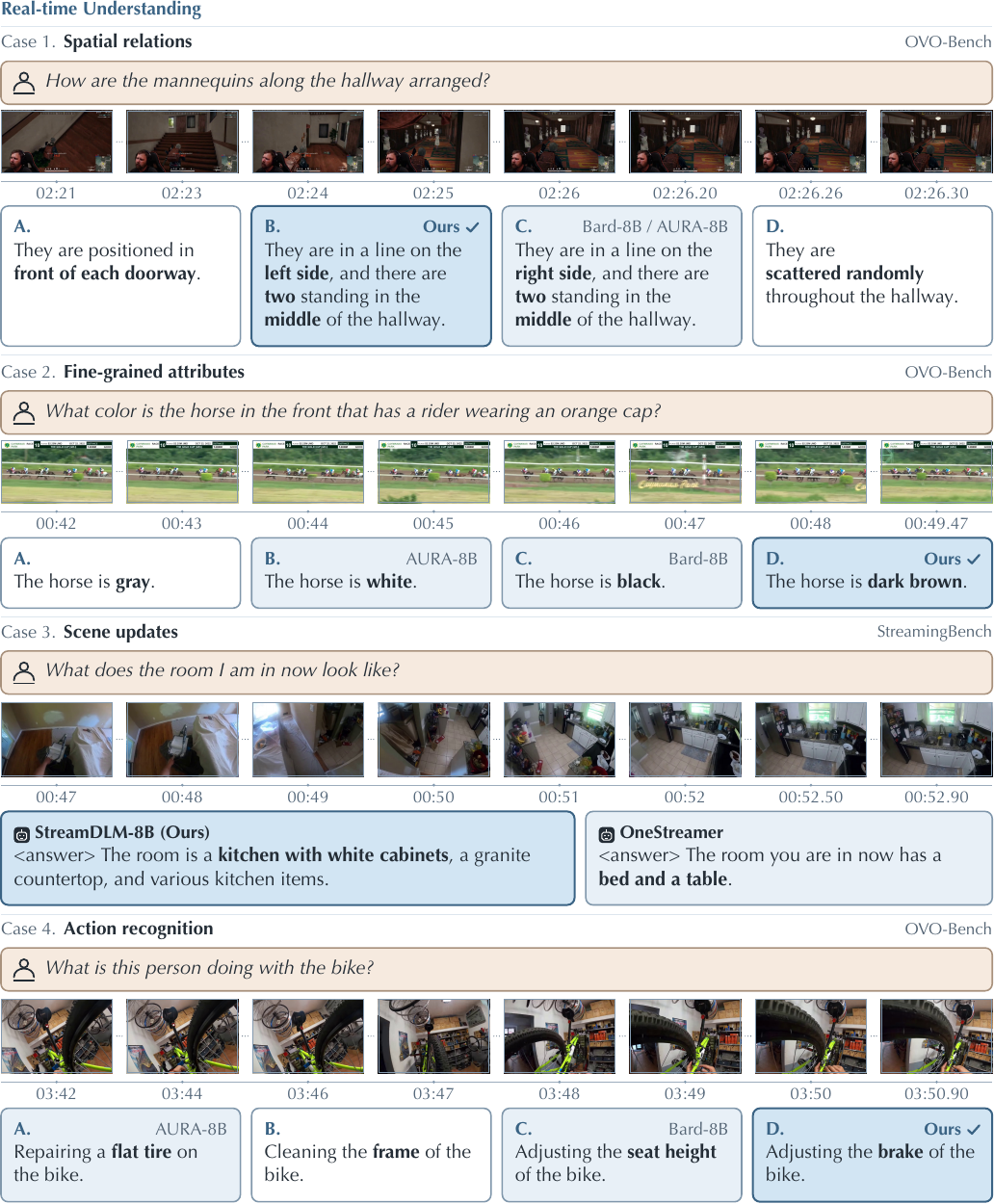}
\caption{\textbf{Real-time understanding.} Understanding spatial relations, visual attributes, and the current scene as the video unfolds.}
\label{fig:qualitative_realtime}
\label{fig:qualitative_realtime_attributes}
\label{fig:qualitative_realtime_scene}
\end{figure*}
\clearpage

\begin{figure*}[p]
\centering
\includegraphics[width=\textwidth]{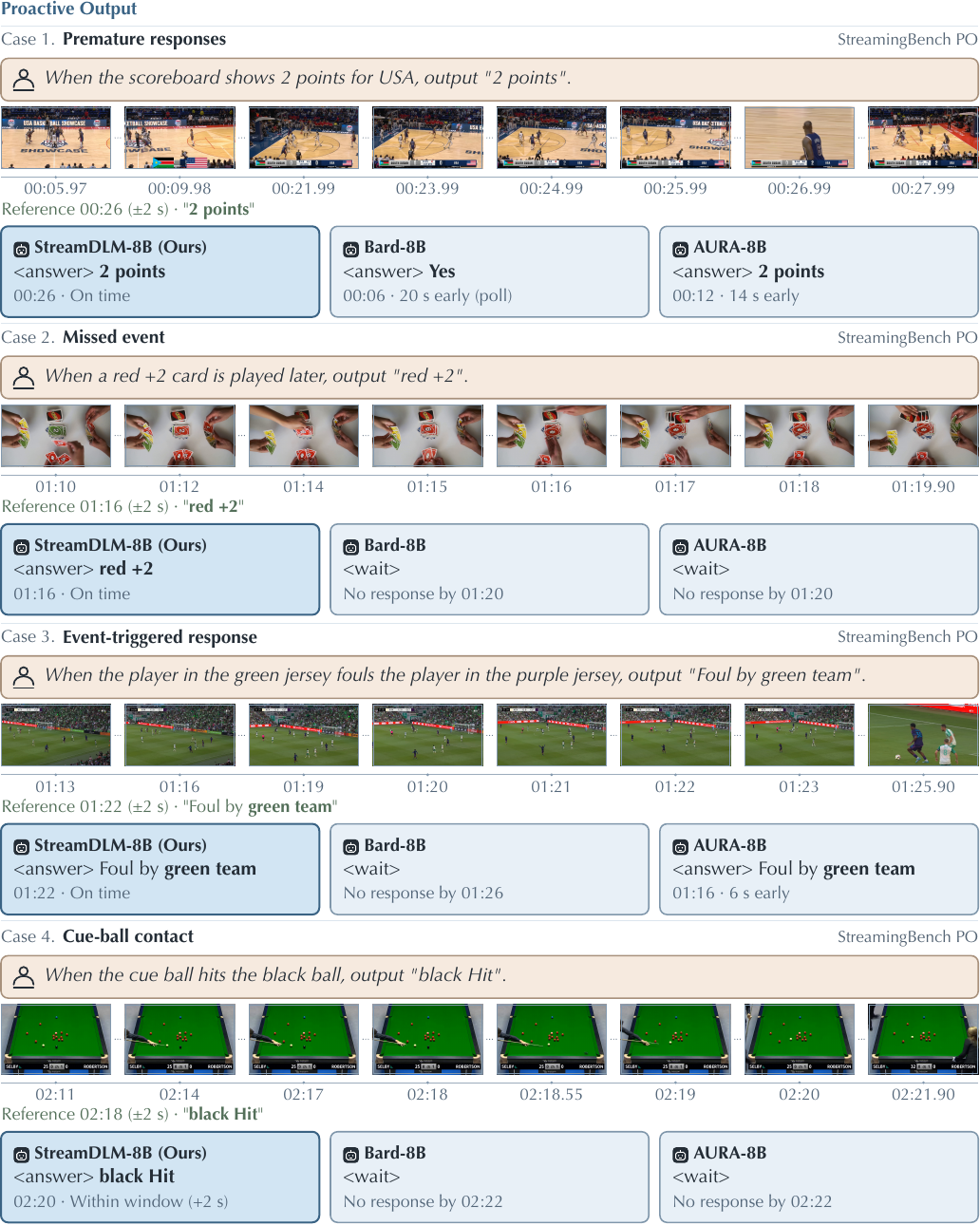}
\caption{\textbf{Proactive output.} Responding when the requested event occurs while remaining silent beforehand.}
\label{fig:qualitative_proactive}
\label{fig:qualitative_proactive_missed}
\label{fig:qualitative_proactive_foul}
\end{figure*}
\clearpage

\begin{figure*}[p]
\centering
\includegraphics[width=\textwidth]{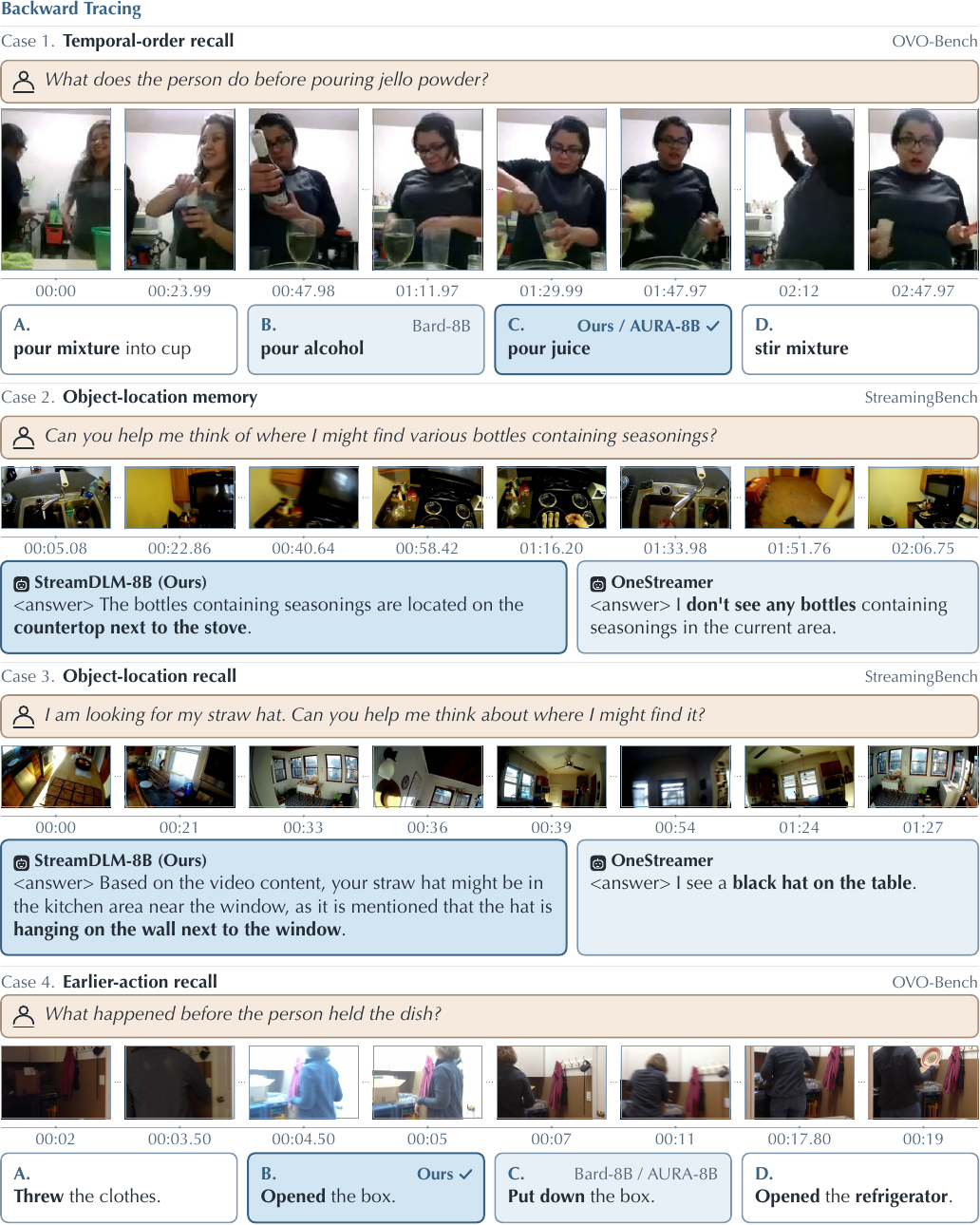}
\caption{\textbf{Backward tracing.} Recalling earlier actions and object locations from the video history.}
\label{fig:qualitative_backward}
\label{fig:qualitative_backward_bottles}
\label{fig:qualitative_backward_hat}
\end{figure*}
\clearpage
\endgroup

\end{document}